\documentclass[manuscript,screen]{acmart}

\AtBeginDocument{%
  \providecommand\BibTeX{{%
    \normalfont B\kern-0.5em{\scshape i\kern-0.25em b}\kern-0.8em\TeX}}}

\setcopyright{acmcopyright}
\copyrightyear{2023}
\acmYear{2023}
\acmDOI{XXXXXXX.XXXXXXX}

\usepackage{algorithm}
\usepackage{algorithmic}
\usepackage{subfigure}
\usepackage{graphicx}
\usepackage{float}
\usepackage{bm}

\begin{document}

\title{Deep Coupling Network For Multivariate Time Series Forecasting}

\author{Kun Yi}
\email{yikun@bit.edu.cn}
\affiliation{%
  \institution{School of Computer Science, Beijing Institute of Technology}
  \streetaddress{no. 5 Zhongguancun South Street}
  \city{Haidian}
  \state{Beijing}
  \country{China}
  \postcode{100081}
}

\author{Qi Zhang}
\email{zhangqi\_cs@tongji.edu.cn}
\affiliation{%
  \institution{School of Electronic and Information Engineering, Tongji University}
  \country{China}}

\author{Hui He}
\email{hehui617@bit.edu.cn}
\affiliation{%
  \institution{School of Medical Technology, Beijing Institute of Technology}
  \country{China}
}

\author{Kaize Shi}
\email{Kaize.Shi@uts.edu.au}
\affiliation{%
 \institution{Faculty of Engineering and Information Technology, University of Technology Sydney}
 \country{Australia}}

\author{Liang Hu}
\email{rainmilk@gmail.com}
\affiliation{%
  \institution{School of Electronic and Information Engineering, Tongji University}
  \country{China}}

\author{Ning An}
\email{ning.g.an@acm.org}
\affiliation{%
  \institution{School of Computer Science and Information Engineering, HeFei University of Technology}
  \country{China}}

\author{Zhendong Niu}
\email{zniu@bit.edu.cn}
\affiliation{%
  \institution{School of Computer Science, Beijing Institute of Technology}
  \country{China}}

  
\renewcommand{\shortauthors}{Kun Yi, et al.}

\begin{abstract}
Multivariate time series (MTS) forecasting is crucial in many real-world applications. To achieve accurate MTS forecasting, it is essential to simultaneously consider both intra- and inter-series relationships among time series data. However, previous work has typically modeled intra- and inter-series relationships separately and has disregarded multi-order interactions present within and between time series data, which can seriously degrade forecasting accuracy. In this paper, we reexamine intra- and inter-series relationships from the perspective of mutual information and accordingly construct a comprehensive relationship learning mechanism tailored to simultaneously capture the intricate multi-order intra- and inter-series couplings. Based on the mechanism, we propose a novel deep coupling network for MTS forecasting, named DeepCN, which consists of a coupling mechanism dedicated to explicitly exploring the multi-order intra- and inter-series relationships among time series data concurrently, a coupled variable representation module aimed at encoding diverse variable patterns, and an inference module facilitating predictions through one forward step. Extensive experiments conducted on seven real-world datasets demonstrate that our proposed DeepCN achieves superior performance compared with the state-of-the-art baselines.
\end{abstract}






\maketitle

\section{Introduction}
Multivariate time series (MTS) forecasting is widely applied in various real-world scenarios, ranging from weather forecasting \cite{Zheng2015}, traffic forecasting \cite{Yu2018}, COVID-19 prediction \cite{Cao2020}, macroeconomics \cite{Li2014}, financial analysis \cite{Binkowski2018}, stock prediction \cite{FengHWLLC19,LiST19}, to decision making \cite{Borovykh2017}, etc. This broad application has attracted increasing enthusiasm and led to very intensive and diversified MTS research in recent years. In particular, numerous deep learning architectures have been developed to address the challenge of the complex intra- and inter-series relationships among time series \cite{Lim_2021,Torres2020}. 

Previous deep learning work on MTS forecasting can be broadly categorized into two main approaches: sequential models~\cite{Zhou2021,autoformer21} and GNN-based models~\cite{Cao2020,Bai2020nips}. Sequential models, including recurrent neural networks (RNNs) \cite{Salinas2019,Rangapuram2018}, convolutional neural networks (CNNs) \cite{Borovykh2017}, temporal convolutional networks (TCNs) \cite{bai2018}, and attention mechanisms \cite{Fan2019}, have been introduced to tackle MTS problems and have achieved good performance, which is attributed to their capability to extract nonlinear intra-series relationships in MTS. However, these models often fall short in effectively capturing inter-series relationships. In contrast, GNN-based models \cite{Cao2020,wu2020connecting} have recently emerged as promising approaches for MTS forecasting with their essential capability to capture the complex inter-series relationships between variables. They predominantly adopt a graph neural network to extract inter-series relationships while incorporating a temporal network to capture intra-series relationships. Unfortunately, the aforementioned methods still exhibit certain limitations in modeling both intra- and inter-series relationships among time series data. 

\begin{figure}
    \centering
    \includegraphics[width=0.9\linewidth]{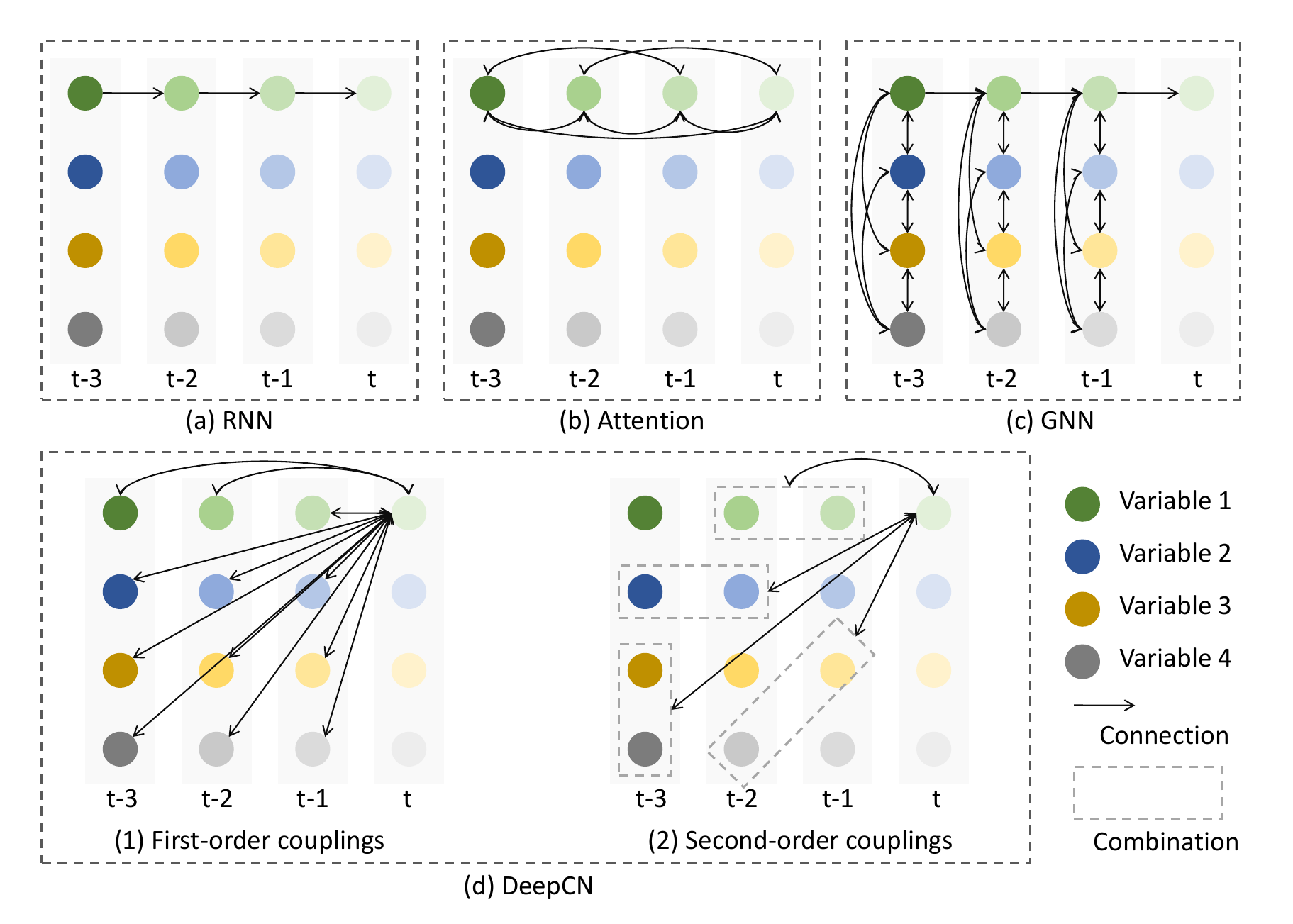}
    \caption{Illustration of modeling the intra- and inter-series relationships. (a) RNN-based models connect two values at consecutive time steps while ignoring the inter-series relationships. (b) Attention-based models directly link variable values across different time steps but do not consider inter-series relationships. (c) GNN-based models construct a graph to model inter-series relationships at each timestamp and then connect values of adjacent time steps for each variable. (d) Our model DeepCN proposes a coupling mechanism to comprehensively capture intra- and inter-series relationships by leveraging the multi-order couplings of various time lags.}
    \label{fig:problem}
\end{figure}

As shown in Fig. \ref{fig:problem}, the maximum length of the signal traversing path of RNN-based models is $\mathcal{O}(L)$, making it challenging for them to effectively learn intra-series relationships between distant positions \cite{LiuYLLLLD22}. Attention-based models \cite{reformer20,Zhou2021,autoformer21} shorten the maximum path to be $\mathcal{O}(1)$ but they ignore inter-series relationships. 
Recently, GNN-based MTS forecasting models \cite{Yu2018,wu2020connecting,zonghanwu2019,Bai2020nips} have emerged. They combine graph networks to represent inter-series relationships while relying on LSTM or GRU to capture intra-series relationships. However, akin to RNN-based models, GNN-based models also encounter limitations in effectively modeling intra-series relationships. Additionally, they are unable to directly connect different variables at different timestamps, thereby limiting their ability to capture abundant inter-series relationships. Moreover, both sequential models and GNN-based models generally model intra- and inter-series relationships separately, which naturally violates the real-world unified intra-/inter-series relationships, and they also overlook the multi-order interactions embodied within time series data. 
Consequently, how to model the complex intra- and inter-series relationships calls for a comprehensive analysis.
Couplings \cite{couplings}, referring to any relationships or interactions that connect two or more variables, include diverse correlations, dependencies, interactions, and hierarchies. Learning such couplings \cite{couplings} can explore more comprehensive representations by revealing and embedding various couplings on complex data \cite{pami_ZhuCY22}, which has demonstrated success in numerous domains, including recommendation \cite{tkde_JianPCLG19,ZhangCSN22}, image source identification \cite{HuangCZPL18}, and financial market analysis \cite{CaoHC15}. Motivated by this, we model the complex intra- and inter-series relationships by exploring multi-order couplings as shown in Fig. \ref{fig:problem}(d). 

In light of the above discussion, in this paper, we theoretically analyze the relationship modeling within time series data and propose a novel deep coupling network (named DeepCN) for multivariate time series forecasting. Concretely, we revisit the relationships from the perspective of mutual information, and then based on the analysis, we first propose a coupling mechanism to comprehensively model intra- and inter-series relationships explicitly and simultaneously via exploring the diverse and hierarchical couplings in MTS data. Subsequently, we leverage a coupled variable representation module to encode the variable relationship representations since different variables exhibit different patterns. Finally, we utilize an inference module to make predictions by one forward step which can avoid error accumulation. We compare the proposed DeepCN with state-of-the-art baselines on seven real-world datasets. Extensive experimental results show the superiority of DeepCN and demonstrate the effectiveness of the proposed neural architecture and coupling mechanism in DeepCN.

Our main contributions are summarized as follows:
\begin{itemize}
    \item Compared with previous models, we theoretically revisit intra- and inter-series relationships through the lens of mutual information and construct our model based on the multi-order couplings which can capture more comprehensive information to enhance relationship representations within time series data.
    \item We design a coupling mechanism to explicitly investigate the complicated intra- and inter-series relationships within time series data simultaneously. The coupling mechanism explores the diverse and hierarchical couplings across various time lags in time series data while maintaining linear computational complexity.
    \item We propose a DeepCN model for MTS forecasting based on the coupling mechanism which captures complex multivariate relationships to address the relationships modeling issues within MTS data.
    \item Extensive experimental results on seven real-world datasets show our DeepCN improves an average improvement of 6.2$\%$ on MAE and 7.0$\%$ on RMSE over state-of-the-art baselines. In addition, more empirical analysis on the coupling mechanism further gains insights into the various performance of different models on different datasets. This analysis provides valuable guidance for effectively handling diverse types of MTS data.
\end{itemize}

The remainder of this paper is structured as follows: In Section \ref{related_work}, we begin with an overview of related work. Subsequently, a description of the problem definition and an introduction of mutual information are given in Section \ref{preliminary}. Moving on, Section \ref{coupling_mechanism} delves into the analysis of the relationships within multivariate time series data, examining them through the lens of mutual information. Following this analysis, in Section \ref{deep_coupling_network}, we propose a coupling mechanism for modeling the relationships, aligning with the insights from Section \ref{coupling_mechanism}, and suggest a deep coupling network tailored for MTS forecasting based on this mechanism. Section \ref{experiments} conducts extensive evaluations on seven real-world datasets to demonstrate the superior performance of our model when compared to state-of-the-art baselines. Finally, we conclude the paper and outline directions for future work in Section \ref{conclusion}.

\section{Related Work} \label{related_work}
Several work has been devoted to multivariate time series forecasting. In this section, we mainly introduce some representative work in MTS forecasting and discuss related methods, including classic models, modern models, representation learning models, and strategies for handling feature interactions.

\subsection{Classic models}
Classic time series forecasting is based on statistical theory and mainly focuses on individual time series \cite{Faloutsos2019}, including autoregressive models (AR) \cite{2015Recursive}, autoregressive integrated moving average (ARIMA) \cite{Asteriou2011}, state space model (SSM) \cite{Hyndman2008}, exponential smoothing (ES) \cite{Abramia}, hidden Markov models (HMM) \cite{Baum1966}, etc. These methods are interpretable, but they require manual feature engineering and usually rely on strong assumptions. Furthermore, the classic methods are linear and mainly focus on intra-series relationships, i.e., ignoring  inter-series correlations.

\subsection{Modern models}
Modern time series forecasting is based on deep learning \cite{Lim2020}, including RNN \cite{HochreiterS97}, CNN \cite{KrizhevskySH12}, GNN \cite{Yu2018,Cao2020}, and attention mechanisms \cite{VaswaniSPUJGKP17}. Compared with the classic methods which can only model linear relationships in data, modern models have shown promising performance on MTS forecasting due to their capability of fitting any complex nonlinear correlations \cite{Lim2020}. Moreover, they can adapt directly to the data without any prior assumptions which provides significant advantages when dealing with little information about the time series \cite{Lara2021}.

\textbf{Sequential Models}. Sequential models, including recurrent neural networks (RNNs) \cite{Salinas2019}, convolutional neural networks (CNNs)~\cite{Lai2018}, temporal convolutional networks (TCNs) \cite{bai2018}, and Transformers \cite{VaswaniSPUJGKP17}, have been adopted to capture intra-series dependencies in deep MTS forecasting models. SFM \cite{ZhangAQ17} decomposes the hidden state of LSTM into multiple frequencies to capture the multi-frequency trading patterns for long and short predictions. LSTNet \cite{Lai2018} leverages CNN to extract short-term local dependency patterns and RNN to discover long-term temporal patterns. IVM \cite{Guo2019} explores the structure of LSTM to infer variable-wise temporal importance. SCINet \cite{liu2022SCINet} proposes a downsample-convolve-interact architecture to enable multi-resolution analysis by expanding the receptive field of TCN. In recent years, to address the overfitting problem due to the small amount of data, hybrid models combining classic time series forecasting and deep learning have been proposed \cite{Lim2020}, such as Exponential Smoothing RNN (ES-RNN) \cite{Smyl2020}, deepAR \cite{Salinas2019} and deepSSM \cite{Rangapuram2018}. Recently, Transformer-based models become prevalent due to their powerful capability of self-attention mechanisms to capture long-term dependencies. TFT \cite{Lim2019} adopts a neural attentive model for interpretable high-performance multi-horizon forecasting. Reformer \cite{reformer20} replaces dot-product attention by using locality-sensitive hashing to address the high complexity. Informer \cite{Zhou2021} proposes a ProbSparse self-attention mechanism and distills operation to handle the challenges of quadratic time complexity and quadratic memory usage in vanilla Transformer. Autoformer \cite{autoformer21} designs a novel decomposition architecture with an Auto-Correlation mechanism. However, these models fall short in capturing inter-series correlations among time series.

\textbf{Matrix factorization Models}. Some previous work apply matrix factorization methods to factorize the relationships between time series into a low-rank matrix, and then perform the forecasting in the low-dimensional latent space. TRMF \cite{Yunips2016} incorporates temporal regularization into matrix factorization formulation. DeepGLO \cite{Sen2019} introduces TCN as a regularization to add non-linear based on TRMF. TLAE \cite{Nguyen2021} advances the global factorization approaches and offers an efficient combination of flexible nonlinear autoencoder mapping and inherent latent temporal dynamics. However, these matrix factorization methods fall short in exploiting the structural dependencies 
among time series \cite{li2019enhancing}.

\textbf{GNN-based Models}. Recently, MTS has embraced GNN \cite{zonghanwu2019,Bai2020nips,wu2020connecting,Yu2018} because of their best capability of modeling inter-series dependencies among time series. DCRNN \cite{LiYS018} leverages bidirectional graph random walk to capture the inter-series dependencies among variables. STGCN \cite{Yu2018} integrates graph convolution and gated temporal convolution for extracting spatial and temporal dependencies. GraphWaveNet \cite{zonghanwu2019} captures spatial and temporal dependencies by combining graph convolution with dilated casual convolution. AGCRN \cite{Bai2020nips} proposes a data-adaptive graph generation module and a node adaptive parameter learning module to enhance the graph convolutional network. MTGNN \cite{wu2020connecting} presents an effective learning method to exploit the inherent inter-series dependencies among time series. 

Among these models, matrix factorization models and GNN-based models can attend to both the intra- and inter-series relationships while sequential models mainly handle the intra-series relationships. Compared with GNN-based models, matrix factorization methods can not model the complex inter-series relationships among time series \cite{li2019enhancing}. However, GNN-based models exploit the relationships through point-wise and pair-wise interactions (as shown in Fig. \ref{fig:problem}(c)) which can not fully express the complex relationships within time series. In this paper, we revisit the relationships (including the intra- and inter-series relationships) within time series data from the perspective of mutual information and propose a deep coupling mechanism to model them.



\subsection{Representation learning for time series forecasting}
Nowadays, there has been a surge of approaches that seek to learn representations to encode the information and dependencies of the time series~\cite{BengioCV13}. The core idea behind time series representation learning is to embed raw time series into a hidden and low-dimensional vector space. In the learning and optimization process for such representations, the complicated dependencies in time series can be encoded and then the learned embeddings can be regarded as feature inputs to downstream machine learning time series tasks.  TS2Vec~\cite{TS2Vec22} utilizes a hierarchical contrastive learning approach on augmented context views to generate robust contextual representations for each timestamp. InfoTS~\cite{InfoTS23} introduces a new contrastive learning method with information-aware augmentations, allowing for the adaptive selection of optimal augmentations to enhance time series representation learning. CoST~\cite{cost22} separates the representation learning and downstream forecasting task and proposes a contrastive learning framework that learns disentangled season-trend representations for time series forecasting tasks.

The key difference that separates time series representation learning methods from common time series-related methods (e.g., sequential models and GNN-based models) is that they treat the learning representations as the target, while other methods are directly designed for finishing time series tasks (e.g., forecasting).

\subsection{Feature interaction} \label{related_feature_interaction}
Feature interactions have been the key to the success of many prediction models \cite{WangFu2017} and are common in many domains, such as Click-Through-Rate (CTR), genetics studies, and environmental effects \cite{Lin2016}. In addition to the linear effects, high-order feature interactions are also important for many complex applications \cite{Lin2016}. Although deep neural networks (DNNs) can learn both low- and high-order feature interactions, they learn implicitly and at bit-wise level \cite{wu2020,LianZZCXS18}. There are three main categories for feature interactions, including aggregation-based method, graph-based method, and combination-based method \cite{wu2020}. Compared with the other two methods, the combination method generates feature interactions explicitly. Wide\&Deep \cite{Cheng0HSCAACCIA16} uses a wide component to generate cross features and takes them as input of deep neural networks. DCN \cite{WangFu2017} leverages a cross network to encode feature interactions explicitly and a neural network to encode implicitly. xDeepFM \cite{LianZZCXS18} uses a Compressed Interaction Network (CIN) to generate feature interactions explicitly. In this paper, motivated by the combination method, we model the relationships within time series data explicitly via a Cartesian Product model \cite{wu2020}. 

\section{Preliminaries}
\label{preliminary}
\subsection{Problem Definition}
We assume a multivariate time series input ${X}_{1:N}^{t-T+1:t}=\{ X_1^{t-T+1:t}, X_2^{t-T+1:t}, ..., X_N^{t-T+1:t}\} \in \mathbb{R}^{N\times T}$ at timestamp $t$ with the number of time series $N$ and the look-back window size $T$, where $X_N^{t-T+1:t} \in \mathbb{R}^{1\times T}$ denotes the $N$-th time series. If $Z$ belongs to $X$ and serves as the target variable, the forecasting task is to predict the next $\tau$ timestamps $\hat{Z}^{t+1:t+\tau} \in \mathbb{R}^{1\times \tau}$ based on the historical $T$ observations ${X}_{1:N}^{t-T+1:t}$. We can formulate the task as follows:
\begin{equation}
    \hat{Z}^{t+1:t+\tau}=F_\Theta \left ({Z}^{t-T+1:t},{X}_{1:N}^{t-T+1:t}  \right )
\end{equation}where $F_\Theta$ represents the prediction function with learnable parameters $\Theta$. To achieve precise multivariate time series forecasting, it is crucial to leverage both the intra-series relationships within $Z^{t-T+1:t}$ and the inter-series relationship between $Z^{t-T+1:t}$ and ${X}_{1:N}^{t-T+1:t}$ \cite{Cao2020, Bai2020nips}. In this paper, we aim to simultaneously capture the multi-order intra- and inter-series relationships and explore their application in multivariate time series forecasting.

\subsection{Mutual Information}
Mutual information is commonly used to describe variable correlations \cite{Wangm2005}, and the definition is as follows:
\begin{equation}
    I(X;Y)=\int_{X}\int_{Y}P(X,Y)log\frac{P(X,Y)}{P(X)P(Y)}
\end{equation}where $X,Y$ are two variables and $P(X),P(Y),P(X,Y)$ are probability distribution and joint probability distribution respectively. 
For multiple variables $X_{1:N}=\{X_1, X_2, ..., X_N\}$ and one variable $Z$,
according to the chain rule of information, the multivariate mutual information between $X_{1:N}$ and $Z$ can be defined as follows:
\begin{equation}\label{mutual}
\begin{split}
    I({X}_{1:N};Z)=\sum_{s\subseteq S}I(\{s\cup Z\}),\left | s \right |\geq 1
\end{split}
\end{equation} where $S=\{{X}_1,{X}_2,...,{X}_N\}$, $s$ is the subset of $S$, and $Z$ is the target variable. This equation sums up the mutual information between the subsets $s$ combined with the target variable $Z$ over all possible subsets of $S$, where each subset 
$s$ contains at least one variable. In this paper, by leveraging the mutual information, we revisit the modeling of intra- and inter-series relationships for multivariate time series forecasting.
To facilitate understanding of the symbols used in this paper, we provide a summary of the key notations in Table \ref{tab:notation}. 
\begin{table}
    \caption{Primary Notations}
    \label{tab:notation}
    \begin{tabular}{l l}
    \toprule
      Symbol   &  Description\\
    \midrule
      ${X}^t$ & the multivariate time series input ${X}$ at timestamp t, ${X}^t \in \mathbb{R}^{N \times T}$\\
      ${X}_N$ & the $N$-th time series of ${X}$, ${X}_N \in \mathbb{R}^{1\times T}$\\
      $\bm{x}$ & the vector $\bm{x}$ reshaped from $X^t$, $\bm{x} \in \mathbb{R}^{NT\times 1}$\\
      N & the number of variables of ${X}$\\
      T & the look-back window size\\
      $\tau$ & the prediction length\\
      Z & the target variable, $Z \in \mathbb{R}^{1 \times T}$\\
      $F_\Theta$ & the prediction function with the learnable parameters $\Theta$ \\
      I & mutual information among time series\\
      S & a set of variables, $S=\{X_1, X_2, ..., X_N\}$\\
      s & the subset of S\\
      $\ell$ & the number of orders of the couplings\\
      $\mathcal{C}^{\ell}$ & the $\ell$-order coupling, $\mathcal{C}^{\ell} \in \mathbb{R}^{NT\times 1}$ \\
      d & the dimension size \\
      $W_{coup}^{\ell}$ & the weight parameters for the $\ell$-order coupling learning, $W_{coup}^{\ell} \in \mathbb{R}^{NT\times 1}$\\
      $b_{coup}^{\ell}$ & the bias parameters for the $\ell$-order coupling learning, $b_{coup}^{\ell} \in \mathbb{R}^{NT\times 1}$\\
      $W_{var}$ & the variable embedding weight matrix, $W_{var} \in \mathbb{R}^{d \times N}$\\
      $W_{h}$ & the weight matrix, $W_{h} \in \mathbb{R}^{NT\times d}$\\
    \bottomrule
    \end{tabular}
    \vspace{-3mm}
\end{table}
\section{Analysis of relationship between time series}\label{coupling_mechanism}
As mentioned in the introduction, the intra- and inter-series relationships among time series are intricate. In this section, we delve deeper into these relationships by revisiting them from the perspective of mutual information.

Specifically, we apply Equation (\ref{mutual}) to multivariate time series analysis scenarios, where we enumerate the subset of $S$ and expand the right side of the equation as follows:
\begin{equation}\label{9}
\begin{split}
        \sum_{s\subseteq S}I(\{s\cup Z\})&=\sum_{i=1}^{N}I({X}_i;Z)
        +\sum_{i=1}^{N}\sum_{j=i+1}^{N}I(\{{X}_i,{X}_j;Z\})\\
        &+\sum_{i=1}^{N}\sum_{j=i+1}^{N}\sum_{k=j+1}^{N}I(\{{X}_i,{X}_j,{X}_k;Z\})+...
\end{split}
\end{equation} where $I({X}_i;Z)$ represents the relationships between ${X}_i$ and $Z$, $I(\{{X}_i,{X}_j;Z\})$ represents the relationships between $Z$ and $\{{X}_i,{X}_j\}$, and $I(\{{X}_i,{X}_j,{X}_k;Z\})$ represents the relationships between $Z$ and $\{{X}_i,{X}_j,{X}_k\}$. 
This expanded equation, denoted as \textbf{multi-order couplings}, illustrates that the relationships between $Z$ and ${X}_{1:N}$ are the summation of mutual information between different combinations of ${X}_{1:N}$ and $Z$.




Moreover, in real-world MTS scenarios, there is often a \textbf{time lag effect} between time-series variables, which is a common phenomenon~\cite{wei1989}. For instance, in a financial portfolio, there might be a time lag influence between two assets, such as the dollar and gold prices, where the value of ${X}^t$ at time $t$ may be influenced by its past values, such as ${X}^{t-1}, {X}^{t-2}, ..., {X}^{t-T+1}$. The presence of time lag effect makes the relationships between variables more intricate. Consequently, to accurately model the relationship between ${X}_1$ and ${X}_2$ at time $t$, we must not only consider the data relation at time $t$ but also take into account the information from previous time steps $t-1,...,t-T+1$. Incorporating this historical context is essential for capturing the complex dynamics and dependencies in the time series data effectively.

Taking into account the above introduction of multi-order couplings and time lag effect, we argue to reconsider the relationship between variable $Z$ and ${X}_{1:N}$ not only at time step $t$ but also at previous time steps $t-1$, ..., $t-T+1$. Thus, we reformulate the relationship as follows:
\begin{equation} \label{10}
\begin{split}
I(Z;{X}_1,{X}_2,...,{X}_N)&= I(Z;{X}_1^{t},{X}_1^{t-1},...,{X}_1^{t-T+1},{X}_2^{t},{X}_2^{t-1},...,{X}_2^{t-T+1},...,{X}_N^{t},{X}_N^{t-1},...,{X}_N^{t-T+1})\\
&= \underbrace{\sum_{i=1}^{N}I({X}_i^t;Z)+\sum_{i=1}^{N}I({X}_i^{t-1};Z)+...+\sum_{i=1}^{N}I({X}_i^{t-T+1};Z)}_{first-order\quad coupling}\\
&+\underbrace{\sum_{i=1}^{N}\sum_{j=i+1}^{N}\sum_{l=1}^{T}\sum_{m=1}^{T}I(\{{X}_i^{t-l+1},{X}_j^{t-m+1};Z\})}_{second-order \quad coupling}\\
&+\underbrace{\sum_{i=1}^{N}\sum_{j=i+1}^{N}\sum_{k=j+1}^{N}\sum_{l=1}^{T}\sum_{m=1}^{T}\sum_{n=1}^{T}I(\{{X}_i^{t-l+1},{X}_j^{t-m+1},{X}_k^{t-n+1};Z\})}_{third-order \quad coupling}+...
\end{split}
\end{equation} where $T$ is the delay time step. Similar to Equation (\ref{9}), the right part of Equation (\ref{10}) is expanded as the summation of mutual information between different combinations of ${X}$ and $Z$. From Equation (\ref{10}), we can conclude that the relationship between $Z$ and ${X}$ can be modeled through multi-order couplings between different time lags of ${X}$ and $Z$. When the target variable $Z$ is one of ${X}_{1:N}$, Equation (\ref{10}) contains both intra-series (e.g., I(${X}_i^t$;${X}_i^{t}$) when Z=${X}_i$) and inter-series (e.g., I($X_i^t$;$X_j^t$) when $i \not= j$) relationships. Therefore, from the mutual information perspective, the intra- and inter-series relationships among time series can be modeled by multi-order couplings of different time lags of ${X}$.

As shown in Fig. \ref{fig:problem}, RNN-based models and attention-based models primarily focus on capturing intra-series relationships. According to Equation (\ref{10}), they are limited in their ability to capture the inter-series dependencies between $X_i$ and $X_j$ ($i\not=j$). Besides, RNN-based models fail to explicitly account for the time lag effect, whereas attention-based models connecting different time lags can capture the effect within time series. That enables attention-based models to capture the couplings between $X_i^{t-l+1}$ and $X_i^{t-m+1}$ ($l\not=m \pm 1$), which RNN-based models can not achieve. Consequently, attention-based models tend to outperform RNN-based models in many cases.
On the other hand, GNN-based models can model both intra-series and inter-series relationships. However, they still lack the capability to explicitly attend to the time lag effect for both intra- and inter-series. Specifically, GNN-based models cannot directly account for the correlations between $X_i^{t-l+1}$ and $X_j^{t-m+1}$ ($l\not=m \pm 1, i\not=j$). This limitation sometimes leads to lower forecasting accuracy compared to attention-based models.

In this paper, to address these shortcomings, we design a comprehensive coupling learning mechanism that explicitly explores diverse and hierarchical couplings among time series, as represented in Equation (\ref{10}). This mechanism aims to capture both intra- and inter-series relationships effectively, while also accounting for the time lag effect, ultimately enhancing the forecasting accuracy in multivariate time series analysis. The proposed model is detailed in Section \ref{model_relationship}. In the experiments, we conduct a parameter sensitivity test on the input length to verify the time lag effect in the relationships between variables. Additionally, we delve into studying multi-order couplings by analyzing the coupling mechanism. For a more comprehensive understanding, more specific analyses are elaborated in Section \ref{experiment_parameter}.

\section{Deep Coupling Network For Multivariate Time Series Forecasting}\label{deep_coupling_network}
In this section, we present a novel deep coupling network for MTS forecasting, called DeepCN. Building upon the insightful analysis in Section \ref{coupling_mechanism}, DeepCN is designed to comprehensively and simultaneously model both the intra- and inter-series relationships among time series, aiming to enhance forecasting performance.
\begin{figure}
    \centering
    \includegraphics[width=1\linewidth]{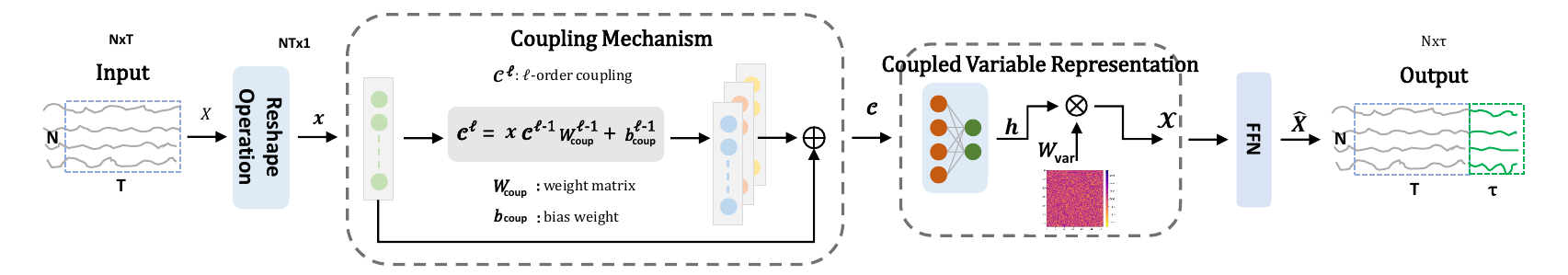}
    \caption{The overview framework of DeepCN, comprising a coupling mechanism, a coupled variable representation module, and an inference module. \textit{Coupling mechanism}: it comprehensively explores the complicated multi-order intra- and inter-series relationships simultaneously among time series data $X$ by a recursive multiplication (see Equation (\ref{crossnet})), and output couplings $\mathcal{C}$. \textit{Coupled variable representation}: it first generates a dense representation $h$ by a fully connected network, and then captures the corresponding variable representations by conducting multiplications between ${h}$ and the weights $W_{var}$ (see Equation (\ref{equation_10})), and outputs $\mathcal{X}$. \textit{Inference}: it makes predictions by a FFN network according to $\mathcal{X}$ and outputs $\hat{X}$.}
    \label{fig:framework}
\end{figure}
\subsection{Overview}
The overall framework of DeepCN is illustrated in Fig. \ref{fig:framework}. Given a multivariate time series input ${X} \in \mathbb{R}^{N \times T}$ where $N$ is the number of variables and $T$ is the look-back window size, first we conduct a reshape operation to transform the input ${X} \in \mathbb{R}^{N \times T}$ to a vector ${\bm{x}} \in \mathbb{R}^{NT\times 1}$. Next, $\bm{x}$ serves as input to the coupling mechanism which is designed to comprehensively explore the complicated intra- and inter-series relationships according to the analysis in Section \ref{coupling_mechanism}. It applies explicit variable crossing of different time lags and outputs the corresponding couplings $\mathcal{C}^{2:\ell}=\{\mathcal{C}^2,...,\mathcal{C}^{\ell}\}$ where $\mathcal{C}^\ell \in \mathbb{R}^{NT\times 1}$ is the $\ell$-order coupling and $\ell$ is the number of order (details see in Subsection \ref{model_relationship}). Then we stack $\mathcal{C}^{2:\ell}$ with ${x}$ and get the multi-order couplings among time series $\mathcal{C}=\{\mathcal{C}^1,...,\mathcal{C}^\ell\}$.
 
After that, $\mathcal{C}$ is fed to a fully connected layer and outputs the dense representation $h \in \mathbb{R}^{\ell \times d}$ where $d$ is the dimension size. Since different variables exhibit different patterns, we initialize a variable embedding matrix $W_{var} \in \mathbb{R}^{N \times d}$ to embed the variable and perform a multiplication between $h$ and $W_{var}$ to represent variable relationships $\mathcal{X} \in \mathbb{R}^{N \times \ell \times d}$. Finally, we utilize a feed-forward network consisting of two fully connected layers and an activation function to predict the next $\tau$ timestamps $\hat{{X}} \in \mathbb{R}^{N \times \tau}$ by one forward step. The above content is explained in detail below.

\subsection{Coupling Mechanism}\label{model_relationship}

A crucial element of our proposed model is the coupling mechanism, which plays a pivotal role in explicitly exploring diverse and hierarchical couplings to capture the complex multi-order intra- and inter-series relationships within time series.
In Section \ref{coupling_mechanism}, we have thoroughly analyzed the characteristics of relationships within time series data. In this subsection, we leverage these insights to design a coupling mechanism that effectively represents relationships based on those identified characteristics, such as multi-order couplings and the time lag effect. This mechanism is integral to our model's ability to model and capture the intricate dependencies within the multivariate time series data.

\begin{figure*}[!t]
    \centering
    \includegraphics[width=0.5\textwidth]{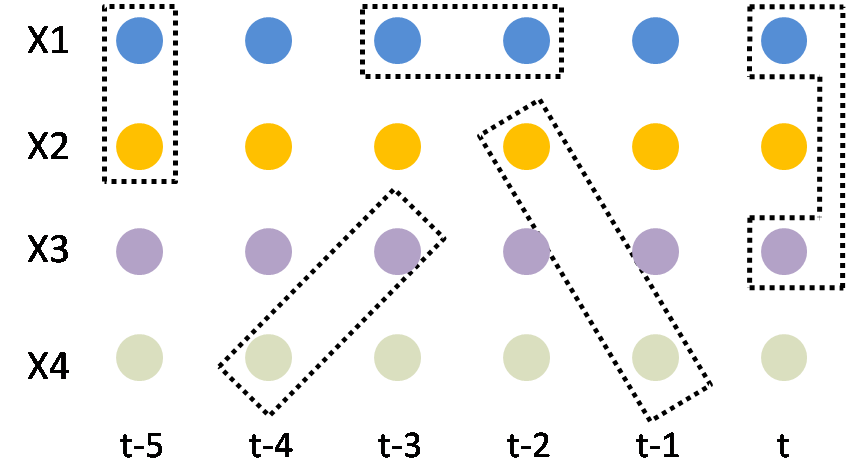}
    \caption{Multi-order couplings diagram. In the figure, we take four variables, five time lags, and second-order as an example. From the figure, there are various combinations of cross-variable and cross-time, and several of them are marked with dotted lines. Moreover, the combinations include both intra-series (e.g., $\{X_{1}^{t-2}X_{1}^{t-3}\}, \{X_{1}^{t-1}X_{1}^{t-3}\}$) and inter-series (e.g., $\{X_{1}^{t}X_{3}^{t}\}, \{X_{2}^{t-2}X_{4}^{t-1}$\}) information.}
    \label{fig:coupling}
\end{figure*}

According to Equation (\ref{10}), the relationships within time series can be expressed through the couplings between different combinations of time lags of time series. 
Intuitively, the multi-order couplings can be modeled by cross feature \cite{WangFu2017}. In light of the content introduced in Subsection \ref{related_feature_interaction}, we understand that combinatorial features are essential in commercial models \cite{LianZZCXS18}. The Cartesian Product model, as a state-of-the-art instance of a combinatorial-based model \cite{wu2020}, is an explicit model. Motivated by this, to explore the $\ell$-order couplings between $Z$ and ${X}_{1:N}^{t-T+1:t}$, we leverage the Cartesian product to illustrate how to calculate it. First, the Cartesian product $Cart_{\ell}$ can be defined as follows:
\begin{equation}\label{cartesian}
\begin{split}
    Cart_\ell = \{(\underbrace{X_i,X_j,...,X_l}_\ell)|X_{i,j,...,l}\in X_{1:N}^{t-T+1:t},\forall i,j,...,l=1,2,...,N\}
\end{split}
\end{equation} where $N$ is the number of variables and $T$ is the length of timestamps. To illustrate this definition more intuitively, we use second-order couplings, five-time delay steps, and four variables as an example in Fig. \ref{fig:coupling}. In the example, there are various combinations of cross variables and cross time, and several of them are marked with dotted lines in the figure. 

As evident from Equation (\ref{cartesian}), $Cart_\ell$ encompasses all possible combinations of $N$ variables from timestamps $t-T+1$ to timestamps $t$. These combinations include both intra-series correlations (e.g., the intra-series correlations for $X_1$ from timestamps $t-T+1$ to timestamps $t$, $X_{1}^{t-T+1}X_{1}^{t-T+2}...X_{1}^{t}$) and inter-series correlations (e.g., the inter-series correlations among $X_1, X_2, ..., X_N$ at timestamps $t$, $X_{1}^{t}X_{2}^{t}...X_{N}^{t}$). 

\begin{figure*}
    \centering
    \includegraphics[width=0.9\textwidth]{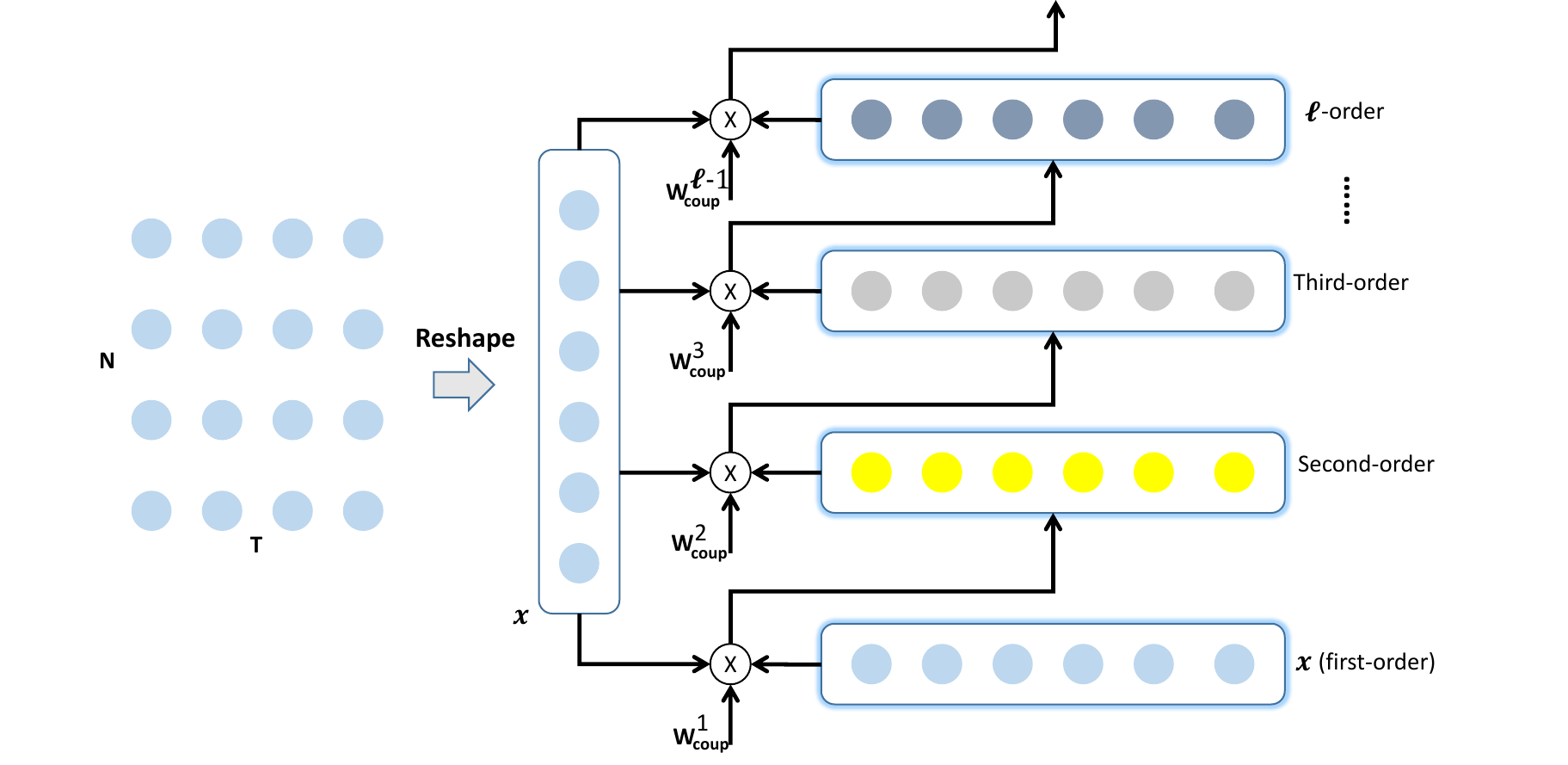}
    \caption{Coupling-based model for relationships between variables. First, we transform the input matrix $X \in \mathbb{R}^{N \times T}$ to an input vector ${x}=\mathbb{R}^{NT\times 1}$ by a reshape operation where N is the number of variables and $T$ is the input length. Then we calculate the different order couplings respectively. $x$ is the first order.}
    \label{fig:coupling_mechanism}
\end{figure*}

Inspired by the cross network \cite{WangFu2017} that has linear complexity, we achieve these combinations explicitly via efficiently learning all types of cross features. First, we reshape the input matrix $X\in \mathbb{R}^{N \times T}$ to a vector ${x} \in \mathbb{R}^{NT\times 1}$. Then we adopt the approach of feature interactions to model the relationships, and we can get the $\ell$-order couplings as follows:
\begin{equation}\label{crossnet}
\begin{split}
    \mathcal{C}^\ell={x}{(\mathcal{C}^{\ell-1})}^TW_{coup}^{\ell-1}+b_{coup}^{\ell-1}, \ell>1\\
    \mathcal{C}^1={x},\ell=1
\end{split}
\end{equation}
where $W_{coup}^{\ell-1} \in \mathbb{R}^{NT \times 1}, b_{coup}^{\ell-1} \in \mathbb{R}^{NT\times 1}$ are weight and bias parameters, respectively. Note that the first-order coupling, namely $\mathcal{C}^1$, is the input ${x}$ itself. The detailed calculation process is shown in Fig. \ref{fig:coupling_mechanism} and the algorithm is described in Algorithm \ref{alg_cm}. According to Equation (\ref{crossnet}), we can find that it can fully explore all types of combinations of $X$. Different values of $\ell$ correspond to different order couplings. For example, when $\ell=2$, $\mathcal{C}^\ell$ corresponds to second-order couplings. Accordingly, we can express the hierarchical couplings $\mathcal{C}$ which are composed of various order couplings as follows:
\begin{equation}
   \mathcal{C} = (\mathcal{C}^1,\mathcal{C}^2,...,\mathcal{C}^{\ell})
\end{equation}
where $\ell$ is the total number of orders. 

\textbf{Complexity Analysis}. Let $\ell$ denote the total number of orders, $N$ denote the number of variables, and $T$ denote the look-back window size. Then the total number of parameters, namely $W_{coup}$ and $b_{coup}$ in Equation (\ref{crossnet}), is 
\begin{equation}\label{equ_complex}
    \ell \times N\times T \times 2.
\end{equation}
From Equation (\ref{equ_complex}), we can conclude that the complexity of the coupling mechanism is linearly proportional to the orders $\ell$, the number of variables $N$, and the input length $T$. In general, $T$ is much smaller than $N$.

\begin{algorithm}
\caption{Coupling Mechanism}
\label{alg_cm}
\begin{algorithmic}[1]
\renewcommand{\algorithmicrequire}{ \textbf{Input:}}
\REQUIRE
multivariate time series ${X}^t \in \mathbb{R}^{N \times T}$
\renewcommand{\algorithmicrequire}{ \textbf{Output:}}
\REQUIRE
the multi-order couplings $\mathcal{C} \in \mathbb{R}^{\ell \times NT}$
\STATE Given the multivariate time series input at timestamp t, $X^t \in \mathbb{R}^{N \times T}$
\STATE Reshape $X$ into a vector ${x} \in \mathbb{R}^{NT \times 1}$
\STATE Initialize the couplings $\mathcal{C}$
\STATE $\mathcal{C}^1 = {x}$
\FOR{$i=2,...,\ell$}
\STATE Initialize weight vector $W_{coup}^{i-1} \in \mathbb{R}^{NT \times 1}$ 
\STATE Initialize bias vector $b_{coup}^{i-1} \in \mathbb{R}^{NT \times 1}$
\STATE $\mathcal{C}^{i} = {x} (\mathcal{C}^{i-1})^T W_{coup}^{i-1}$ + $b_{coup}^{i-1}$
\ENDFOR
\STATE $\mathcal{C}=\{\mathcal{C}^{1},...,\mathcal{C}^{\ell}\}$
\STATE \textbf{return} $\mathcal{C}$
\end{algorithmic}
\end{algorithm}
\subsection{Coupled Variable Representation Module}
The resultant output $\mathcal{C}$ from the coupling mechanism captures a comprehensive representation of both intra- and inter-series dependencies among time series, including first-order coupling, second-order coupling, and beyond. However, given different variables may exhibit distinct patterns, we employ a coupled variable representation module to learn specific representations of the relationships between variables. The module helps to accommodate the diverse nature of MTS data and enables the model to better understand and represent the unique dependencies embodied in MTS.

To start, the output $\mathcal{C} \in \mathbb{R}^{\ell \times NT}$ from the coupling mechanism is inputted into a fully connected network, generating a dense representation $h \in \mathbb{R}^{\ell \times d}$, where $d$ represents the dimension size. Next, we initialize a variable embedding weight matrix $W_{var} \in \mathbb{R}^{d \times N}$ and perform multiplication with the dense representation $h$, resulting in an output $\mathcal{X} \in \mathbb{R}^{N \times \ell \times d}$. The calculation can be described as follows:
\begin{equation}\label{equation_10}
\begin{split}
    h = \mathcal{C}W_h+b_h,\\
    \mathcal{X} = h \times W_{var}
\end{split}
\end{equation} where $W_h \in \mathbb{R}^{NT \times d}$ and $b_h \in \mathbb{R}^{\ell \times d}$ is the weight matrix and bias matrix, respectively, $W_{var} \in \mathbb{R}^{d \times N}$ is the variable embedding weight matrix.

\subsection{Inference Module}
The inference module, which consists of a $ReLU$ activation function and two linear networks, is utilized to make predictions based on the output of the coupled variable representation $\mathcal{X}$. To prevent error accumulation, we generate all prediction values with a single forward step, which has proven to be more efficient and stable compared to step-by-step prediction (as demonstrated in Section \ref{experiments}). The formulation of the inference module is as follows:
\begin{equation}
\label{equ:ffn}
    \hat{{X}}=\sigma(\mathcal{X}{W}_1+{b}_1){W}_2+{b}_2
\end{equation}
where $\mathcal{X} \in \mathbb{R}^{N \times \ell \times d}$ is the output of coupled variable representation module, $\sigma$ is the activation function, ${W}_1 \in \mathbb{R}^{(\ell*d)\times d_h}, {W}_2 \in \mathbb{R}^{d_h \times \tau}$ are the weights, and ${b}_1 \in \mathbb{R}^{d_h}$, ${b}_2 \in \mathbb{R}^{\tau}$ are the biases, and $d_h$ is the inner-layer dimension size. The final loss can be formulated as follows:
\begin{equation}
    \mathcal{L}(\hat{X},X;\Theta)  = \sum_{t}\left \| \hat{X}^{t+1:t+\tau}-X^{t+1:t+\tau}\right \|_2^2
\end{equation} where $\tau$ is the prediction length, $\Theta$ is the learnable parameters, $\hat{X}$ is the prediction values and $X$ is the ground truth. 

\section{Experiments}\label{experiments}
In this section, we conduct a comprehensive evaluation of our proposed DeepCN model, assessing its prediction accuracy and computational efficiency. Additionally, we provide in-depth analyses of DeepCN, including a study of the coupling mechanism, an ablation study, a model analysis, and a visualization analysis. These evaluations aim to offer a thorough understanding of the capabilities and performance of our DeepCN model.

\subsection{Datasets}
We perform extensive experiments on seven real-world datasets, and a summary of these datasets is presented in Table \ref{tab:datasets}. To ensure uniformity, all datasets are normalized using min-max normalization. Regarding data splitting, we divide all datasets, except for the COVID-19 dataset, into training, validation, and test sets, maintaining a chronological order with a ratio of 7:2:1. Given the relatively small sample size of the COVID-19 dataset, we have adjusted the data splitting ratio to 6:2:2.

\textbf{Solar}\footnote{\url{https://www.nrel.gov/grid/solar-power-data.html}}: This data set is about solar power collected by NREL (National Renewable Energy Laboratory). We use the usage of a state as the data set which contains 593 points. The data has been collected from 2006/01/01 to 2016/12/31 with the sampling interval of every 1 hour. 

\textbf{Wiki\footnote{\url{https://www.kaggle.com/c/web-traffic-time-series-forecasting/data}}}: The data set contains a number of daily views of different Wikipedia articles which have been collected from 2015/7/1 to 2016/12/31. It consists of approximately 145k time series and we choose 2k of them as our experimental data set due to limited computing resources.

\textbf{Traffic\footnote{\url{https://archive.ics.uci.edu/ml/datasets/PEMS-SF}}}: The data set contains a number of hourly traffic data of 963 San Francisco freeway car lanes which are collected from 2015/01/01 with the sampling interval of every 1 hour.

\textbf{ECG\footnote{\url{http://www.timeseriesclassification.com/description.php?Dataset=ECG5000}}}: The data set is about Electrocardiogram(ECG) from the UCR time-series classification archive \cite{dau2018}. It contains 140 nodes and each node has a length of 5000.

\textbf{COVID-19\footnote{\url{https://github.com/CSSEGISandData/COVID-19}}}: The dataset is about COVID-19 hospitalization in the U.S. state of California (CA) from 01/02/2020 to 31/12/2020 provided by the Johns Hopkins University with the sampling interval of every day.

\textbf{Electricity\footnote{\url{https://archive.ics.uci.edu/ml/datasets/ElectricityLoadDiagrams20112014}}}: This dataset contains the electricity consumption of 370 clients and is collected since 2011/01/01. The data sampling interval is every 15 minutes.

\textbf{METR-LA\footnote{\url{https://github.com/liyaguang/DCRNN}}}: This dataset contains traffic information collected from loop detectors in the highway of Los Angeles County from 01/03/2012 to 30/06/2012. It contains 207 sensors and the data sampling interval is every 5 minutes.

\begin{table}[]
    \centering
    \caption{Summary of Experimental Datasets}
    \resizebox{\textwidth}{!}{
    \begin{tabular}{c c c c c c c c c c c}
    \toprule
    Datasets & Solar & Wiki & Traffic & ECG  & COVID-19 & Electricity & METR-LA\\
    \midrule
      Samples & 3650 & 803& 10560& 5000  & 335 & 140211&  34272\\
      Variables & 592 & 5000& 963& 140  & 55 & 370&  207\\
      Granularity & 1hour & 1day& 1hour& - & 1day & 15min&  5min\\
      Start time & 01/01/2006 & 01/07/2015 & 01/01/2015 & - & 01/02/2020 &01/01/2011 & 01/03/2012\\
    \bottomrule
    \end{tabular}}
    \label{tab:datasets}
\end{table}

\subsection{Baselines}
We compare our DeepCN with twenty-three representative and state-of-the-art (SOTA) models, including a classic model, DNN-based models, matrix factorization models, GNN-based models, Transformer-based models, and representation learning models. 

\textbf{Classic Model}.
\begin{itemize}
    \item \textbf{VAR}\footnote{\url{https://www.statsmodels.org}}: VAR \cite{Vector1993} is a classic linear autoregressive model. We use the Statsmodels library which is a Python package that provides statistical computations to realize the VAR.
\end{itemize}

\textbf{DNN-based models}.
\begin{itemize}
    \item \textbf{DA-RNN}\footnote{\url{https://github.com/Zhenye-Na/DA-RNN}}: DA-RNN~\cite{DARNN17} proposes a novel dual-stage attention-based recurrent neural network, consisting of an encoder with an input attention mechanism and a decoder with a temporal attention mechanism. We follow the recommended settings.
    \item \textbf{LSTNet}\footnote{\url{https://github.com/laiguokun/LSTNet}}: LSTNet \cite{Lai2018} uses a CNN to capture inter-variable relationships and an RNN to discover long-term patterns. In our experiment, we use the settings where the number of CNN hidden units is 100, the kernel size of the CNN layers is 4, the dropout is 0.2, the RNN hidden units is 100, the number of RNN hidden layers is 1, the learning rate is 0.001 and the optimizer is Adam.
    \item \textbf{TCN}\footnote{\url{https://github.com/locuslab/TCN}}: TCN \cite{bai2018} is a causal convolution model for regression prediction. We utilize the same configuration as the polyphonic music task exampled in the open source code where the dropout is 0.25, the kernel size is 5, the number of hidden units is 150, the number of levels is 4 and the optimizer is Adam.
    \item \textbf{SFM}\footnote{\url{https://github.com/z331565360/State-Frequency-Memory-stock-prediction}}: On the basis of the LSTM model, SFM \cite{ZhangAQ17} introduces a series of different frequency components in the cell states. We use the default settings as the authors recommended where the learning rate is 0.01, the frequency dimension is 10, the hidden dimension is 10 and the optimizer is RMSProp. 
    \item \textbf{DLinear}\footnote{\url{https://github.com/cure-lab/LTSF-Linear}}: DLinear~\cite{DLinear_2023} proposes a set of embarrassingly simple one-layer linear models to learn temporal relationships between input and output sequences. We use it as our baseline and follow the recommended settings as experimental configurations. 
    \item \textbf{N-BEATS}\footnote{\url{https://github.com/ElementAI/N-BEATS}}: N-BEATS~\cite{oreshkin2019n} utilizes stacked MLP layers together with doubly residual learning to process the input data to iteratively forecast the future. We follow the recommended settings as experimental configurations with 10 stacks, 4 layers, and 256 layer sizes.
    \item \textbf{N-HiTS}\footnote{\url{https://github.com/cchallu/n-hits}}: N-HiTS~\cite{NHiTS23} incorporates novel hierarchical interpolation and multi-rate data sampling techniques to improve forecast accuracy and reduce computational complexity. We use the recommended settings with 3 blocks, 9 layers, hidden size being 512, and learning rate being $0.001$.
\end{itemize}

\textbf{Matrix factorization models}.
\begin{itemize}
    \item \textbf{DeepGLO}\footnote{\url{https://github.com/rajatsen91/deepglo}}: DeepGLO \cite{Sen2019} models the relationships among variables by matrix factorization and employs a temporal convolution neural network to introduce non-linear relationships. We use the default setting as our experimental settings for the Wiki, Electricity, and Traffic datasets. For the COVID-19 dataset, the vertical and horizontal batch size is set to 64, the rank of the global model is set to 64, the number of channels is set to [32, 32, 32, 1], and the period is set to 7.
\end{itemize}

\textbf{GNN-based models}.
\begin{itemize}
    \item \textbf{StemGNN}\footnote{\url{https://github.com/microsoft/StemGNN}}: StemGNN \cite{Cao2020} leverages GFT and DFT to capture dependencies among variables in the frequency domain. We use the default setting of StemGNN as our experiment setting where the optimizer is RMSProp, the learning rate is 0.0001, the stacked layers is 5, and the dropout rate is 0.5.
    \item \textbf{MTGNN}\footnote{\url{https://github.com/nnzhan/MTGNN}}: MTGNN \cite{wu2020connecting} proposes an effective method to exploit the inherent dependency relationships among multiple time series. Because the experimental datasets have no static features, we set the parameter load\_static\_feature to false. We construct the graph by the adaptive adjacency matrix and add the graph convolution layer. Regarding other parameters, we adopt the default settings.
    \item \textbf{GraphWaveNet}\footnote{\url{https://github.com/nnzhan/Graph-WaveNet}}: GraphWaveNet  \cite{zonghanwu2019} introduces an adaptive dependency matrix learning to capture the hidden spatial dependency. Since our datasets have no prior defined graph structures, we use only adaptive adjacent matrix. We add a graph convolutional layer and randomly initialize the adjacent matrix. We adopt the default setting as our experimental settings where the learning rate is 0.001, the dropout is 0.3, the number of epochs is 50, and the optimizer is Adam.
    \item \textbf{AGCRN}\footnote{\url{https://github.com/LeiBAI/AGCRN}}: AGCRN \cite{Bai2020nips} proposes a data-adaptive graph generation module for discovering spatial correlations from data. We use the default settings as our experimental settings where the embedding dimension is 10, the learning rate is 0.003, and the optimizer is Adam.
    
    \item \textbf{TAMP-S2GCNets}\footnote{\url{https://www.dropbox.com/sh/n0ajd5l0tdeyb80/AABGn-ejfV1YtRwjf_L0AOsNa?dl=0}}: TAMP-S2GCNets~\cite{tampsgcnets2022} explores the utility of MP to enhance knowledge representation mechanisms within the time-aware DL paradigm. TAMP-S2GCNets require a pre-defined graph topology and we use the California State topology provided by the source code as input. We adopt the recommended settings as the experimental configuration for COVID-19.

    \item \textbf{DCRNN}\footnote{\url{https://github.com/liyaguang/DCRNN}}: DCRNN~\cite{LiYS018} uses bidirectional graph random walk to model spatial dependency and recurrent neural network to capture the temporal dynamics. We use the recommended configuration as our experimental settings with a batch size of 64, the learning rate being $0.01$, the input dimension being 2, and the optimizer of Adam. DCRNN requires a pre-defined graph structure and we use the adjacency matrix as the pre-defined structure provided by the METR-LA dataset.

    \item \textbf{STGCN}\footnote{\url{https://github.com/VeritasYin/STGCN_IJCAI-18}}: STGCN~\cite{Yu2018} integrates graph convolution and gated temporal convolution through spatial-temporal convolutional blocks. We follow the recommended settings as our experimental configuration where the batch size is 50, the learning rate is $0.001$ and the optimizer is Adam. STGCN requires a pre-defined graph structure and we leverage the adjacency matrix as the pre-defined structure provided by the METR-LA dataset.
\end{itemize}

\textbf{Transformer-based models}.
\begin{itemize}
    \item \textbf{Informer}\footnote{\url{https://github.com/zhouhaoyi/Informer2020}}: Informer \cite{Zhou2021} leverages an efficient self-attention mechanism to encode the dependencies among variables. We use the default settings as our experimental settings where the dropout is 0.05, the number of encoder layers is 2, the number of decoder layers is 1, the learning rate is 0.0001, and the optimizer is Adam.
    \item \textbf{Reformer}\footnote{\url{https://github.com/thuml/Autoformer}}: Reformer \cite{reformer20} combines the modeling capacity of a Transformer with an architecture that can be executed efficiently on long sequences and with small memory use. We follow the recommended settings.
    \item \textbf{Autoformer}\footnote{\url{https://github.com/thuml/Autoformer}}: Autoformer \cite{autoformer21} proposes a decomposition architecture by embedding the series decomposition block as an inner operator, which can progressively aggregate the long-term trend part from intermediate prediction. We follow the recommended settings with 2 encoder layers and 1 decoder layer.
    \item \textbf{PatchTST}\footnote{\url{https://github.com/PatchTST}}: PatchTST \cite{PatchTST2023} proposes an effective design of Transformer-based models for time series forecasting tasks by introducing two key components: patching and channel-independent structure.  We use it as our forecasting baseline and adhere to the recommended settings as the experimental configuration.
\end{itemize}

\textbf{Representation learning-based models}.
\begin{itemize}
    \item \textbf{TS2Vec}\footnote{\url{https://github.com/yuezhihan/ts2vec}}: TS2Vec~\cite{TS2Vec22} is a universal framework for learning representations of time series in an arbitrary semantic level. Following the recommended settings, we set the representation dimension to 320, the hidden dimension of the encoder to 64, the number of hidden residual blocks to 10, and the learning rate to 0.001.
    \item \textbf{InfoTS}\footnote{\url{https://github.com/chengw07/InfoTS}}: InfoTS~\cite{InfoTS23} proposes a new contrastive learning approach with information-aware augmentations that adaptively selects optimal augmentations for time series representation learning. We set the representation dimension to 320, the hidden dimension to 60, and the meta learning rate to 0.01.
    \item \textbf{CoST}\footnote{\url{https://github.com/salesforce/CoST}}: CoST~\cite{cost22} separates the representation learning and downstream forecasting task and proposes a contrastive learning framework that learns disentangled season-trend representations for time series forecasting tasks. We set the representation dimension to 320, the learning rate to 0.001, and the batch size to 32. 
\end{itemize}

\subsection{Experimental Setup}\label{experiment_setup}
We perform our experiments on a single NVIDIA RTX 3080 10G GPU. Our code is implemented by Python 3.6 with PyTorch 1.9. Our model is optimized with RMSprop optimizer and the learning rate is $1e^{-5}$. Across all datasets, batch size is set to 32. The number of epochs is 50. For the Traffic, Electricity, METR-LA, and COVID-19 datasets, the dimension size $d$ is set to 512. For the Wiki, Solar, and ECG datasets, the dimension size $d$ is set to 1024. For the COVID-19, ECG, and Electricity datasets, the number of orders $\ell$ is set to 2. For the Wiki and Traffic datasets, $\ell$ is set to 4. For the Solar and METR-LA datasets, $\ell$ is set to 3. In the inference module, the hidden size is 1024 and the activation function is $ReLU$.

We use MAE and RMSE as metrics.
Specifically, given the groudtruth $X^{t+1:t+\tau}_{1:N}\in\mathbb{R}^{N\times\tau}$ and the predictions $\hat{X}^{t+1:t+\tau}_{1:N} \in\mathbb{R}^{N\times\tau}$ for future $\tau$ steps at timestamp $t$, the metrics are defined as follows:
\begin{equation}
    MAE=\frac{1}{\tau N}\sum_{i=1}^{N}\sum_{j=1}^{\tau}\left | X_{i}^j-\hat{X}_{i}^j \right |
\end{equation}
\begin{equation}
   RMSE=\sqrt{\frac{1}{\tau N}\sum_{i=1}^{N}\sum_{j=1}^{\tau}\left (X_{i}^j-\hat{X}_{i}^j\right )^2}.
\end{equation}

\begin{table*}[!t]
    \centering
    \caption{Single step forecasting error results (MAE and RMSE) of DeepCN and other baseline models on five datasets with the prediction length being 12 and the input length being 12.}
    \resizebox{\textwidth}{!}{
    \begin{tabular}{l|c c|c c|c c|c c|c c|c c}
    \toprule
        Dataset & \multicolumn{2}{c|}{Solar} &  \multicolumn{2}{c|}{Wiki} & \multicolumn{2}{c|}{Traffic}&   \multicolumn{2}{c|}{ECG} & \multicolumn{2}{c|}{COVID-19} & \multicolumn{2}{c}{Electricity} \\
        Metrics & MAE & RMSE& MAE & RMSE & MAE & RMSE & MAE & RMSE & MAE & RMSE & MAE & RMSE\\
        \midrule
         VAR & 0.184& 0.234 & 0.057& 0.094 & 0.535 &1.133 & 0.120& 0.170 & 0.226 & 0.326 &0.101& 0.163 \\
         SFM & 0.161 & 0.283 & 0.081 & 0.156 & 0.029& 0.044& 0.095 & 0.135 &0.205 &0.308 &0.086 &0.129 \\
         LSTNet  & 0.148 &0.200 & 0.054 &0.090 & 0.026 &0.057 & 0.079 &0.115 & 0.248  & 0.305 & 0.075 &0.138 \\
         TCN  & 0.176 & 0.222 & 0.094 & 0.142 & 0.052 & 0.067 & 0.078 & 0.107 & 0.317 & 0.354 &  0.057 & 0.083\\
         DA-RNN & 0.156 &0.214& 0.060& 0.104& 0.036& 0.055& 0.074& 0.108& 0.212& 0.278& 0.068& 0.114 \\
         DeepGLO  & 0.178 &0.400 & 0.110& 0.113 & 0.025 & 0.037 & 0.110 &0.163 & 0.191 & 0.253 & 0.090 & 0.131\\
         Reformer & 0.234 & 0.292 &0.047 & 0.083& 0.029&0.042 & 0.062& 0.090&\underline{0.182} & {0.249} & 0.078 & 0.129 \\
         Informer  & 0.151 & 0.199 & 0.051 & 0.086 & 0.020 &0.033 & 0.056 &0.085  & 0.200 & 0.259 & 0.070 & 0.119\\
         Autoformer & 0.150& {0.193} &0.069 & 0.103 &0.029 & 0.043 & {0.055} &0.081 &0.189 & \underline{0.241} &{0.056} & {0.083} \\
         FEDformer & \underline{0.139}&\underline{0.182} &0.068&0.098 &0.025&0.038 & 0.055 & 0.080 & 0.190&0.259 & \underline{0.055}& \underline{0.081}\\
         GraphWaveNet & 0.183 & 0.238 & 0.061 & 0.105 & \underline{0.013} & 0.034  & 0.093 & 0.142 & 0.201 & 0.255 &0.094 & 0.140\\
         StemGNN & 0.176 &0.222 & 0.190 &0.255 & 0.080 &0.135 & 0.100 &0.130 & 0.421 & 0.508 &0.070 & 0.101  \\
         MTGNN & 0.151 & 0.207 & 0.101 & 0.140 & \underline{0.013} & \underline{0.030} & 0.096 & 0.145 & 0.394 & 0.488 & 0.077 & 0.113  \\
         AGCRN & {0.143} &0.214 & \underline{0.044} & \underline{0.079} & 0.084& 0.166 & {0.055} & {0.080} & 0.254 & 0.309 & 0.074 & 0.116 \\
         N-BEATS & \underline{0.139}	&\underline{0.182}&	0.047&	0.087&	0.016&	0.031&	0.056&	0.088 &	0.192&	0.292&	0.057&	0.085\\
         N-HITS & 0.129&	0.174&	0.046&	0.083&	0.018&	0.035&	0.055&	0.086 &	0.185&	0.283&	0.060&	0.089\\
         DLinear & 0.257 & 0.313 &0.182 &0.219 &0.046 &0.064 &0.133 &0.176 & 0.306& 0.382& 0.058& 0.092\\
         PatchTST &0.140 &0.234& 0.050& 0.094& 0.016& 0.037& 0.054& 0.080& 0.230& 0.289& 0.058& 0.094 \\
         TS2Vec & 0.141 & 0.184 & 0.045 & 0.080 & 0.021& 0.034& {0.053} & {0.078} & 0.376 & 0.439& 0.105 & 0.156 \\
         InfoTS & 0.142 &0.186& 0.052& 0.094& 0.027& 0.049& \underline{0.052}& \underline{0.077}& 0.330& 0.412& 0.068& 0.104 \\
         \midrule
         \textbf{DeepCN(ours)} & \textbf{0.132} &\textbf{0.172} & \textbf{0.042} &\textbf{0.078} & \textbf{0.011} &\textbf{0.022}  & \textbf{0.051} &\textbf{0.076}  & \textbf{0.170}& \textbf{0.232} & \textbf{0.053}&\textbf{0.078}\\
         Improvement &5.0\% &5.5\%  & 4.5\% &1.3\% &  15.4\% &26.7\% & 1.9\% &1.3\%  & 6.6\% &  3.7\% & 3.6\% & 3.7\%\\ 
         \bottomrule
    \end{tabular}}
    \label{tab:result}
\end{table*}

\subsection{Results}
\textbf{Single-step forecasting.} We compare our model DeepCN against the other baseline models on seven real-world datasets with the input length being 12 and the prediction length being 12. The primary results are summarized in Table \ref{tab:result}. As indicated in the table, DeepCN consistently demonstrates strong performance across all datasets. On the Solar dataset, DeepCN improves 5.0$\%$ on MAE and 5.5$\%$ on RMSE. On the Wiki dataset, it improves 4.5$\%$ on MAE and 1.3$\%$ on RMSE. On the ECG dataset, it improves 1.9$\%$ on MAE and 1.3$\%$ on RMSE. On the COVID-19 dataset, it improves 6.6$\%$ on MAE and 3.7$\%$ on RMSE. On the Electricity dataset, it improves 3.6$\%$ on MAE and 3.7$\%$ on RMSE. Especially, on the Traffic dataset, DeepCN improves 15.4$\%$ on MAE and 26.7$\%$ on RMSE compared with the best baseline. The reason why DeepCN performs exceptionally well on the Traffic dataset is because of the strong coupling in the traffic data. For example, adjacent nodes affect each other, and adjacent areas also affect each other. If a road is jammed, it inevitably affects other roads. We conduct more experiments on the couplings of the Traffic dataset, and the result is shown in Fig. \ref{fig:traffic_study}(a). From the figure, we conclude there are multi-order couplings in the Traffic dataset. This also sheds light on the superior accuracy of DeepCN, as it incorporates the consideration of multi-order couplings.

Among these baseline models, transformer-based models and GNN-based models have achieved more competitive performance than other DNN-based models. As shown in Fig. \ref{fig:problem}, DNN-based models can not attend to the time lag effect directly and ignore the inter-series relationships. The two inherent defects affect their ability to capture the dependencies among time series. Moreover, GNN-based models perform well on Solar, Wiki, Traffic, and ECG datasets while transformer-based models achieve good results only on COVID-19 and ECG datasets. This is because compared with transformer-based models, GNN-based models consider inter-series relationships. The results also indicate the significance of inter-series relationships in MTS forecasting, especially in tight coupling scenarios like traffic forecasting.

\begin{table*}[!t]
    \centering
    \caption{Multi-step forecasting error results (MAE and RMSE) on ECG dataset. We compare DeepCN with the other six baseline models on the ECG dataset when the prediction length is 3, 6, 9, and 12, respectively.}
    \resizebox{\textwidth}{!}{
    \begin{tabular}{l|c c|c c|c c|c c}
    \toprule
       Horizon & \multicolumn{2}{c|}{3} &  \multicolumn{2}{c|}{6} & \multicolumn{2}{c|}{9}  & \multicolumn{2}{c}{12}\\
       Metrics & MAE &RMSE &  MAE& RMSE & MAE &RMSE & MAE& RMSE\\
       \midrule
         LSTNet & 0.085 &0.178& 0.128 &0.202& 0.128 &0.202& 0.128 &0.203\\
         DeepGLO & 0.083 &0.142&0.093 &0.161& 0.107 &0.157& 0.110 &0.163\\
         Informer & \underline{0.055} & \underline{0.083}& \underline{0.055} & \underline{0.084}& \underline{0.057} & \underline{0.086}& \underline{0.056} & \underline{0.085}\\
         GraphWaveNet & 0.090 &0.139 & 0.092 &0.141& 0.095 &0.145& 0.097 &0.149\\
         StemGNN & 0.090 &0.130 & 0.100 &0.130& 0.090 &0.129& 0.100 &0.130\\
         MTGNN & 0.092 &0.140 & 0.094 &0.140& 0.095 &0.144& 0.096 &0.145\\
         \midrule
         \textbf{DeepCN(ours)} & \textbf{0.050} &\textbf{0.076}& \textbf{0.050} &\textbf{0.076}& \textbf{0.051} &\textbf{0.076}&\textbf{0.051} &\textbf{0.076}\\
         Improvement &9.1\% &8.4\%& 9.1\% &9.5\%& 10.5\% &11.6\%& 8.9\% &10.6\% \\ 
         \bottomrule
    \end{tabular}}
    \label{tab:ECG_result}
\end{table*}

\begin{table*}[!t]
    \centering
    \caption{Multi-step forecasting error results (MAE and RMSE) on Traffic dataset. We compare DeepCN with the other seven baseline models on the Traffic dataset when the prediction length is 3, 6, 9, and 12, respectively.}
    \resizebox{\textwidth}{!}{
    \begin{tabular}{l|c c|c c|c c|c c}
    \toprule
       Horizon & \multicolumn{2}{c|}{3} &  \multicolumn{2}{c|}{6} & \multicolumn{2}{c|}{9}  & \multicolumn{2}{c}{12}\\
       Metrics & MAE &RMSE &  MAE& RMSE & MAE &RMSE & MAE& RMSE\\
        \midrule
         VAR & 0.047 &0.076  & 0.095 &0.150 & 0.182 &0.319 & 0.535 &1.133\\
         LSTNet & {0.016} &0.038 & {0.019} &0.045 & {0.023} &0.051 & 0.026 &0.057\\
         DeepGLO & 0.020 & {0.036} & 0.022 &{0.036} & 0.024 &{0.038} & {0.025} & {0.037} \\
         Informer & 0.019 &0.031 & 0.020 &0.032 & 0.020 &0.032  & 0.020 &0.033\\
         GraphWaveNet & \underline{0.011} &0.027 & 0.013 &0.031& 0.013 &0.030& 0.013 &0.034\\
         StemGNN & 0.050 &0.093 & 0.070 &0.121 & 0.090 &0.144 & 0.080 &0.135\\
         MTGNN & \underline{0.011} & \underline{0.026} & \underline{0.012} & \underline{0.027}& \underline{0.012} &\underline{0.028}& \underline{0.013} &\underline{0.030}\\
         \midrule
         \textbf{DeepCN(ours)} & \textbf{0.009} &\textbf{0.020} & \textbf{0.010} &\textbf{0.021} & \textbf{0.011} &\textbf{0.021}  & \textbf{0.011} &\textbf{0.022}\\
         Improvement &18.2\% &23.1\% & 16.7\% &22.2\% & 8.3\% &25.0\% & 15.4\% &26.7\% \\  
         \bottomrule
    \end{tabular}}
    \label{tab:traffic_result}
\end{table*}

\begin{table*}[ht]
    \centering
    \caption{Multi-step forecasting error results (MAE and RMSE) on Wiki dataset. We compare DeepCN with other five baseline models on Traffic dataset when prediction length is 3, 6, 9, and 12, respectively.}
    \resizebox{\textwidth}{!}{
    \begin{tabular}{l|c c|c c|c c|c c}
    \toprule
       Length & \multicolumn{2}{c|}{3} &  \multicolumn{2}{c|}{6} & \multicolumn{2}{c|}{9}  & \multicolumn{2}{c}{12}\\
       Metrics & MAE &RMSE&  MAE& RMSE&MAE &RMSE& MAE& RMSE\\
       \midrule
         GraphWaveNet & 0.061 &0.105 & 0.061 &0.105  & 0.061 & 0.105 & 0.061 &0.104 \\
         StemGNN & 0.157 &0.236 & 0.159 &0.233 & 0.232 &0.311 & 0.220 &0.306 \\
         AGCRN & \underline{0.043} & \underline{0.077} & \underline{0.044} & \underline{0.078} & \underline{0.045} & \underline{0.079} & \underline{0.044} & \underline{0.079} \\
         MTGNN & 0.102 & 0.141 & 0.091 &0.133 & 0.074 &0.120 & 0.101 & 0.140 \\
         Informer & 0.053 &0.089 & 0.054 &0.090 & 0.059 &0.095 & 0.059 &0.095 \\
         \midrule
         \textbf{DeepCN(ours)} & \textbf{0.041} &\textbf{0.076} & \textbf{0.042} &\textbf{0.076} & \textbf{0.042} &\textbf{0.077} & \textbf{0.042} &\textbf{0.078}\\
         Improvement &4.7\% &1.3\% & 4.5\% &2.6\% & 6.7\% &2.5\% & 4.5\% &1.3\% \\ 
         \bottomrule
    \end{tabular}}
    \label{tab:wiki_result}
\end{table*}

\begin{table*}[ht]
    \centering
    \caption{Multi-step forecasting error results (MAE and RMSE) on METR-LA dataset. We compare DeepCN with other eight baseline models on METR-LA dataset when prediction length is 3, 6, 9, and 12, respectively.}
    \resizebox{\textwidth}{!}{
    \begin{tabular}{l|c c|c c|c c|c c}
    \toprule
       Horizon & \multicolumn{2}{c|}{3} &  \multicolumn{2}{c|}{6} & \multicolumn{2}{c|}{9}  & \multicolumn{2}{c}{12}\\
       Metrics & MAE &RMSE&  MAE& RMSE&MAE &RMSE & MAE& RMSE\\
       \midrule
         DCRNN & 0.160 & 0.204 & 0.191 & 0.243& 0.216 & 0.269 & 0.241 & 0.291\\
         STGCN & 0.058 & 0.133 &0.080 &0.177 &0.102 & 0.209 &0.128 &0.238 \\
         GraphWaveNet & 0.180 &0.366 & 0.184 &0.375 & 0.196 & 0.382 & 0.202 &0.386 \\
         MTGNN & 0.135 & 0.294 & 0.144 &0.307 & 0.149 &0.328& 0.153 & 0.316 \\
         StemGNN & \underline{0.052} & \underline{0.115} & \underline{0.069} & \underline{0.141} & \underline{0.080} & 0.162 & \underline{0.093} & 0.175 \\
         AGCRN & 0.062 & 0.131  & 0.086 & 0.165 & 0.099 & 0.188 & 0.109 & 0.204 \\
         Informer  & 0.076 &0.141 & 0.088 &0.163  & 0.096 &0.178 & 0.100 &0.190\\
         CoST  &  0.064 &0.118 & 0.077 & \underline{0.141} & 0.088 &\textbf{0.159} & 0.097 & \underline{0.171} \\
         \midrule
         \textbf{DeepCN(ours)} & \textbf{0.051} &\textbf{0.114} & \textbf{0.068} &\textbf{0.140} & \textbf{0.078} &\textbf{0.159} & \textbf{0.090} &\textbf{0.168}\\
         Improvement &1.9\% &0.9\% & 1.4\% &0.7\% & 2.5\% &-& 3.2\% &1.8\%\\ 
         \bottomrule
    \end{tabular}}
    \label{tab:metr_result}
\end{table*}

\textbf{Multi-step forecasting.} In order to further evaluate the accuracy of DeepCN under different prediction lengths, we conduct more experiments on multi-step forecasting, including 3, 6, 9, and 12 steps. We perform experiments on the ECG, Traffic, Wiki, and METR-LA datasets, respectively. For the ECG dataset, the input length of all models is set to 12. The coupling number $\ell$ of DeepCN is set to 2. As you can see from the results in Table \ref{tab:ECG_result}, DeepCN achieves good performance in multi-step forecasting tasks as it improves an average of 9.4$\%$ on MAE and 10.0$\%$ on RMSE. Among the baselines, Informer performs better than others since it can model the intra-series dependencies directly. For the Traffic dataset, the input length of all models is also set to 12. The coupling number $\ell$ of DeepCN is set to 5. The results are shown in Table \ref{tab:traffic_result} and we can find that DeepCN improves an average of 14.7$\%$ on MAE and 24.3$\%$ on RMSE. Among the baselines, MTGNN shows good performance because it has good capability to capture the inter-series dependencies. For the Wiki dataset, we choose GNN-based models and Transformer-based model Informer as the baseline models, the input length is set to 12 and the number of orders $\ell$ is set to 4. The results in Table \ref{tab:wiki_result} demonstrate that our proposed model improves an average of 5.1$\%$ on MAE and 1.9$\%$ on RMSE. Among the baselines, AGCRN achieves competitive results since it constructs the graph structure (i.e., the inter-series dependencies) adaptively from time series data. For the METR-LA dataset, we compare DeepCN with the GNN-based, Transformer-based, and representation learning based models as our baselines, and the results are shown in Table \ref{tab:metr_result}. The table showcases that on average we improve $2.3\%$ on MAE and $1.1\%$ on RMSE. Among these models, StemGNN achieves competitive performance because it explicitly model both intra- and inter-series dependencies. However, it is limited to simultaneously capturing intra-/inter-series dependencies. CoST learns disentangled seasonal-trend representations for time series forecasting via contrastive learning and obtains competitive results. But, our model still outperforms CoST. Because, compared with CoST, our model not only can learn the dynamic intra-series representations, but also capture the discriminative inter-series representations.

\begin{figure*}[!t]
    \centering
    \subfigure[Analysis on MAE]
    {
    \centering
    \includegraphics[width=2in]{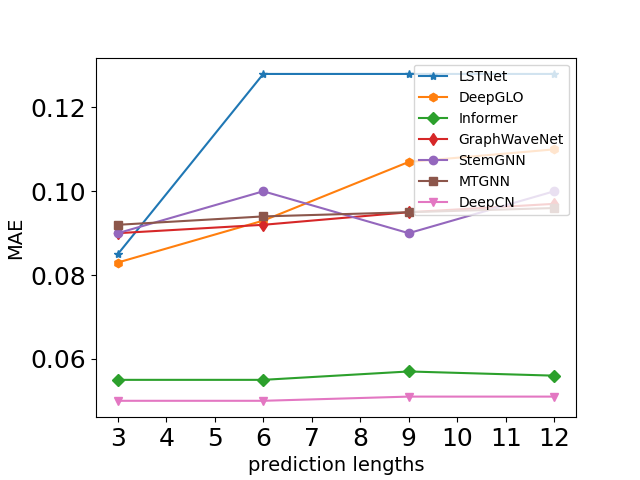}
    }
    \subfigure[Analysis on RMSE]
    {
    \centering
    \includegraphics[width=2in]{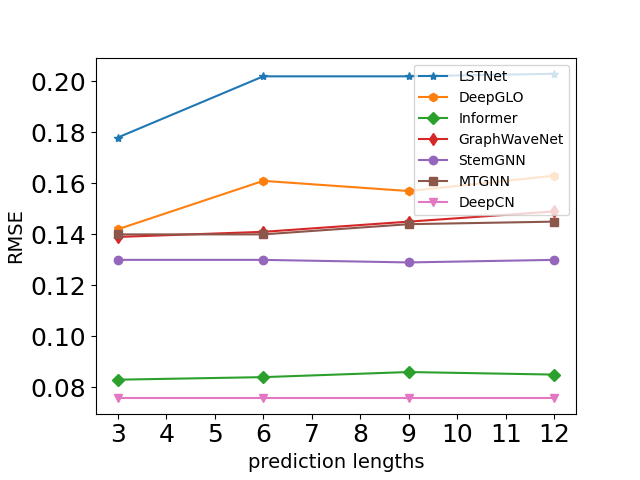}
    }
    \caption{Multi-step forecasting error result analysis (MAE and RMSE) on the ECG dataset. We compare the changing curve of DeepCN with the other five baseline models on the ECG dataset under different prediction lengths (3, 6, 9, 12), respectively.}
    \label{fig:ecg_multistep}
\end{figure*}

\begin{figure*}[!t]
    \centering
    \subfigure[Analysis on MAE]
    {
    \centering
    \includegraphics[width=2in]{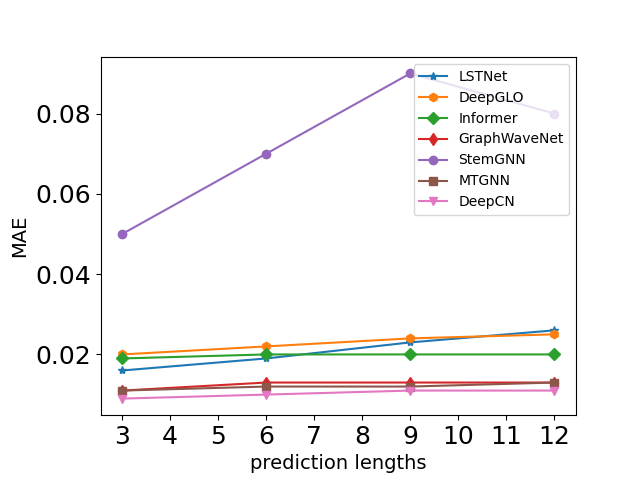}
    }
    \subfigure[Analysis on RMSE]
    {
    \centering
    \includegraphics[width=2in]{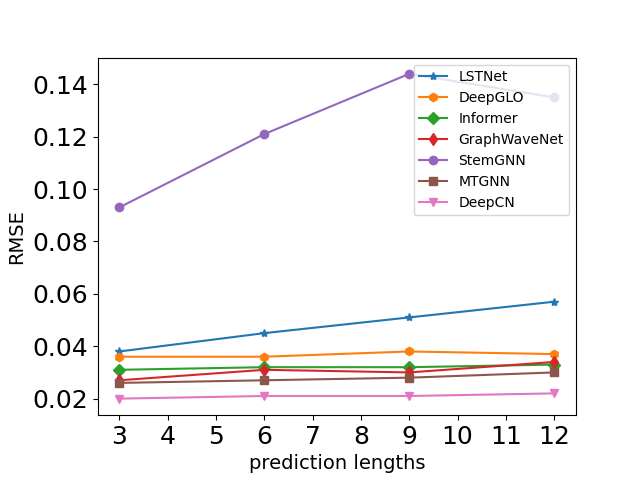}
    }
    \caption{Multi-step forecasting error result analysis (MAE and RMSE) on the Traffic dataset. We compare the changing curve of our model with the other four baseline models on the Traffic dataset under different prediction lengths (3, 6, 9, 12), respectively. We delete the VAR model because its value is too large to affect the display.}
    \label{fig:traffic_multistep}
\end{figure*}

Moreover, Fig. \ref{fig:ecg_multistep} and Fig. \ref{fig:traffic_multistep} exhibit the changing curve of the accuracy under different steps on the ECG dataset and Traffic dataset, respectively. From Fig. \ref{fig:ecg_multistep}, we can find that as the steps increase, the accuracy rate of a classic model (VAR) decreases. It also shows that the accuracy of Informer is closer to ours, but LSTNet's performance is not quite ideal on the ECG dataset. Fig. \ref{fig:traffic_multistep} shows that the accuracy of GNN-based models is closer to ours, whereas StemGNN presents poor accuracy on the Traffic dataset. In addition, from Fig. \ref{fig:ecg_multistep} and Fig. \ref{fig:traffic_multistep}, we can observe that the accuracy of our proposed model DeepCN is stable when prediction length increases because DeepCN utilizes one forward step to make a prediction which can avoid error accumulations. Informer also makes predictions by one forward step and its accuracy shows stable while StemGNN adopting a rolling strategy has fluctuations in performance. 

\begin{figure*}[!t]
    \centering
    \subfigure
    {
    \centering
    \includegraphics[width=2in]{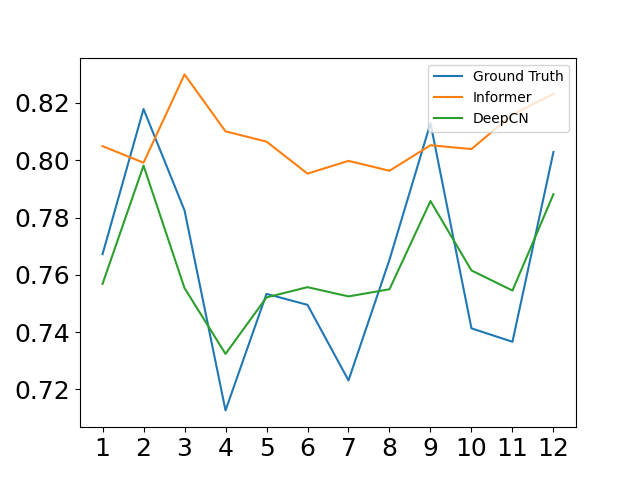}
    }
    \subfigure
    {
    \centering
    \includegraphics[width=2in]{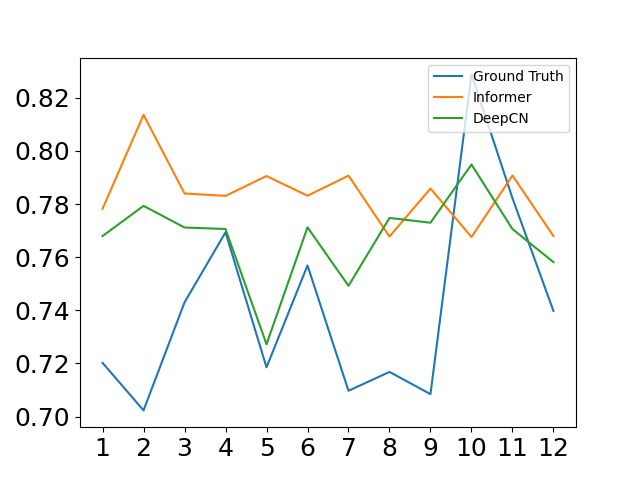}
    }
    
    \subfigure
    {
    \centering
    \includegraphics[width=2in]{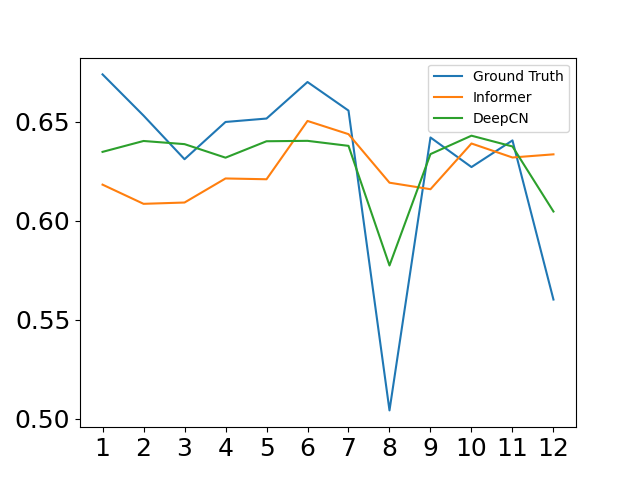}
    }
    \subfigure
    {
    \centering
    \includegraphics[width=2in]{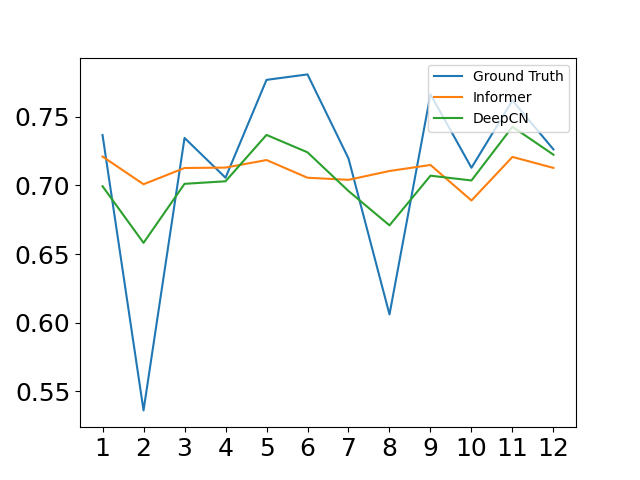}
    }
    \caption{The predicts (steps=12) of DeepCN and Informer on ECG dataset. The three different color curves stand for slices of the ground truth, Informer, and DeepCN, respectively. Four nodes were randomly selected from 140 in the dataset as observation variables to compare the ground truth with the prediction values of Informer and DeepCN. The above four subgraphs correspond to a node, respectively.}
    \label{fig:curve_multistep}
\end{figure*}

Fig. \ref{fig:curve_multistep} shows the comparison of our proposed model and Informer between the predicted values and ground truth when the prediction length is 12. We choose Informer as the comparative baseline because it performs better compared with other models (as shown in Table \ref{tab:ECG_result}). We randomly select four variables from the ECG dataset. Each subgraph corresponds to one variable. The x-coordinate represents the prediction time step and the y-coordinate represents the value. As shown in Fig. \ref{fig:curve_multistep}, the prediction performance of DeepCN is better than that of Informer and it is able to fit the curve of ground truth except for some sudden changes.

\subsection{Analysis}
\begin{figure*}[!t]
    \centering
    \subfigure[accuracy analysis]
    {
    \centering
    \includegraphics[width=0.31\linewidth]{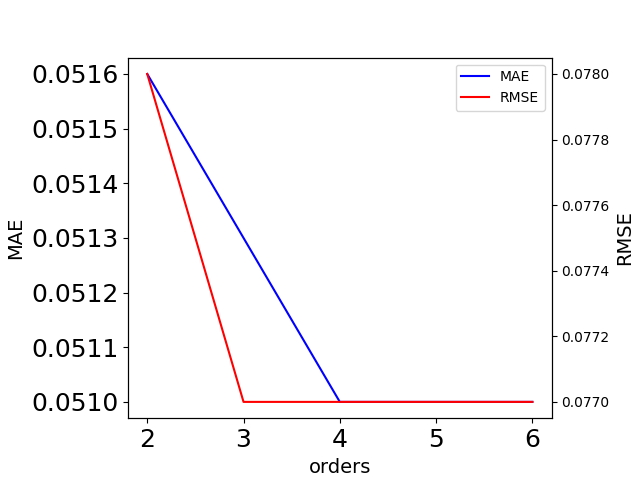}
    }
    \subfigure[efficiency analysis]
    {
    \centering
    \includegraphics[width=0.31\linewidth]{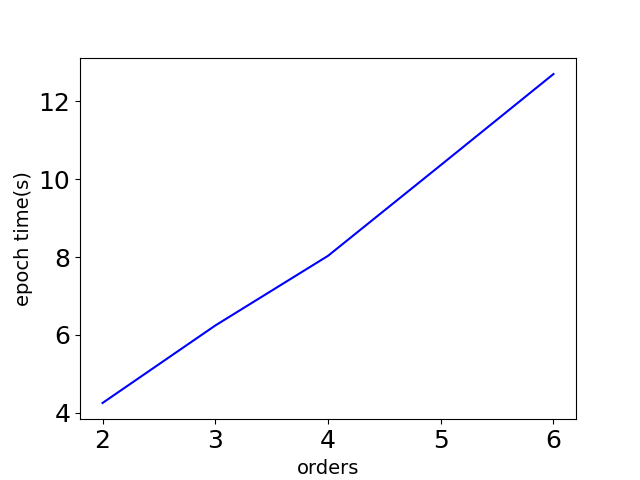}
    }
    \subfigure[train loss analysis]
    {
    \centering
    \includegraphics[width=0.31\linewidth]{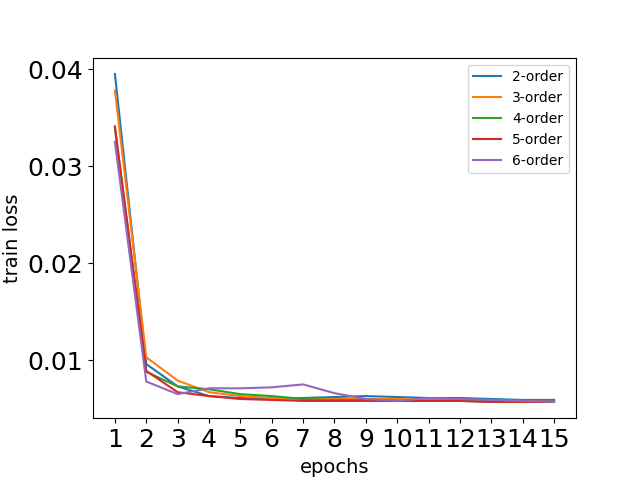}
    }
    \caption{Study of coupling mechanism on ECG dataset. We analyze the accuracy, efficiency, and training loss of our model on the ECG dataset under different orders, respectively. (a) The error result (MAE and RMSE) under different orders. (b) The average training time under different orders. (c) The training loss in different epochs under different orders.}
    \label{fig:solar_study}
\end{figure*}

\begin{figure*}[!t]
    \centering
    \subfigure[accuracy analysis]
    {
    \centering
    \includegraphics[width=0.31\linewidth]{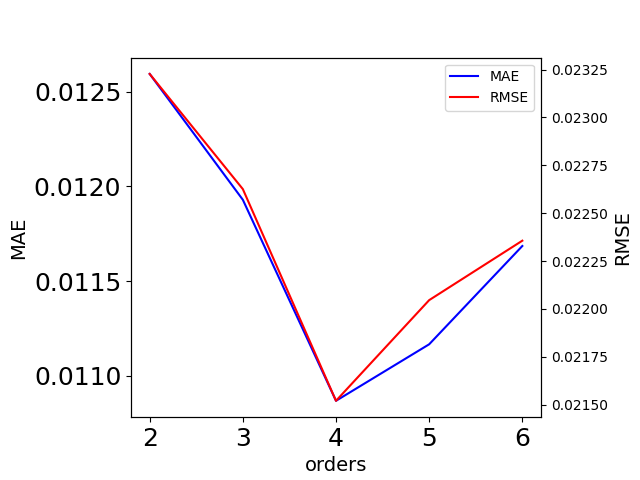}
    }
    \subfigure[efficiency analysis]
    {
    \centering
    \includegraphics[width=0.31\linewidth]{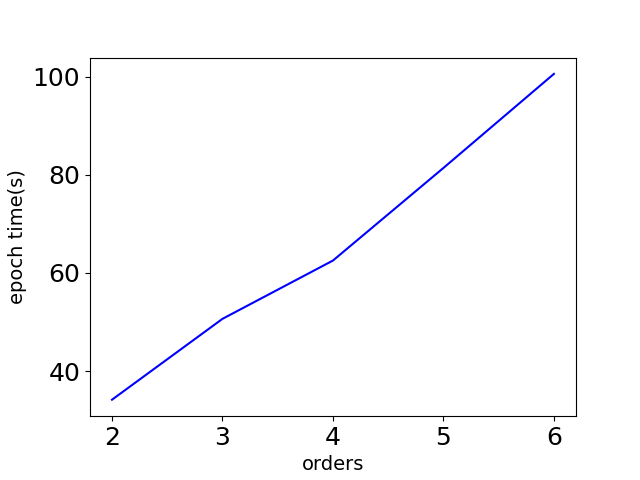}
    }
    \subfigure[train loss analysis]
    {
    \centering
    \includegraphics[width=0.31\linewidth]{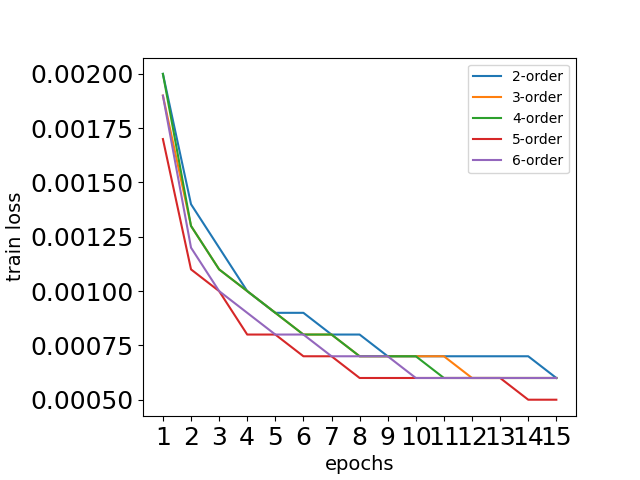}
    }
    \caption{Study of coupling mechanism on the Traffic dataset. We analyze the accuracy, efficiency, and training loss of our model on the Traffic dataset under different orders, respectively. (a) The error result (MAE and RMSE) under different orders. (b) The average training time under different orders. (c) The training loss in different epochs under different orders. }
    \label{fig:traffic_study}
\end{figure*}

\textbf{Study of the Coupling Mechanism.}\label{experiment_coupling} This part mainly addresses how the coupling mechanism affects the accuracy and efficiency of our proposed model. We conduct experiments by imposing a different number of coupling orders $\ell$ (e.g., second-order coupling, third-order coupling, fourth-order coupling, and so on) on the Traffic and ECG datasets, respectively. We separately analyze the relationship between training time and the number of coupling orders, and the relationship between error results (MAE and RMSE) and the number of coupling orders. The result on the ECG dataset is shown in Fig. \ref{fig:solar_study} and the result on the Traffic data is shown in Fig. \ref{fig:traffic_study}. From Fig. \ref{fig:solar_study}, we can see that: 1) As the number of orders $\ell$ increases, the training time becomes longer, but it has little effect on the convergence speed of training. 2) When the order $\ell$ is greater than 2, the accuracy is almost unchanged because the data on the ECG dataset has weak correlations among time series. Namely, the data on the ECG dataset does not exhibit high-order coupling characteristics. From Fig. \ref{fig:traffic_study}, we can find that: 1) With the number of orders $\ell$ increasing, the training time becomes longer and the convergence rate of training does not change much. 2) The accuracy rate increases first which shows that higher-order couplings exist on the Traffic dataset. Then with $\ell$ increasing, the performance worsens since higher order leads to overfitting. From Fig. \ref{fig:solar_study}(a) and Fig. \ref{fig:traffic_study}(a), we can conclude that for the strong inter-series correlation time series dataset (e.g., Traffic dataset), it is necessary to model the high-order couplings, while for the weak inter-series correlation dataset (e.g., ECG dataset), modeling for the high-order couplings does not bring improvements in effectiveness. This also explains that sequential models achieve generally better results than GNN-based models on the ECG dataset, while on the Traffic dataset, GNN-based models perform better than other models.

\textbf{Ablation Study}. 
To further analyze the different orders of couplings, we conduct more experiments on the Traffic dataset and ECG dataset to evaluate the effectiveness of different orders. We set the input length and prediction length to 12, and the number of orders $\ell$ is set to 4. Other experimental settings are the same as introduced in Section \ref{experiment_setup}. We evaluate the effectiveness of each order of coupling by masking the corresponding order. The results are shown in Table \ref{tab:ablation}. In the table, \textbf{w/o coupling} means that the model is without the coupling mechanism. Namely, the model is only composed of the first-order coupling and the fully-connected network. \textbf{w/o 1st} represents the coupling mechanism without the first-order coupling and we mask the first-order coupling from the relationships $\mathcal{C}$. In a similar fashion, \textbf{w/o 2st}, \textbf{w/o 3st}, and \textbf{w/o 4st} represents the coupling mechanism without the second-order, third order and fourth-order couplings, respectively. From the table, we can find that: 1) For the Traffic dataset, each order of coupling is indispensable. 2) For the ECG dataset, compared with other orders of couplings, the first-order coupling is more important. It shows that the high-order couplings should be considered in the strong correlation data while the low-order coupling is enough for the relationship modeling in the weak correlation data.

\begin{table*}[]
    \centering
    \caption{Error results (MAE and RMSE) under different orders of couplings}
    \resizebox{\textwidth}{!}{
    \begin{tabular}{l c c c c c c c}
    \toprule
    Dataset & Metrics & DeepCN & w/o coupling & w/o 1st & w/o 2st &w/o 3st & w/o 4st\\
    \midrule
       {Traffic}&MAE  &  \textbf{0.011} &0.013 & 0.012 & 0.012& 0.012 &0.012 \\
               &RMSE  & \textbf{0.022} & 0.024& 0.024 & 0.023 &0.023 &0.022 \\
    \midrule
       {ECG}&MAE& \textbf{0.050} & 0.051&0.055 & 0.051 & 0.051 & 0.051\\
                &RMSE & \textbf{0.076}& 0.077& 0.081 & 0.077 & 0.077 & 0.077\\
    \bottomrule
    \end{tabular}}
    \label{tab:ablation}
\end{table*}

\textbf{Efficiency Analysis}. To evaluate the efficiency of DeepCN, we compare the training time and parameter counts of our proposed model with GNN-based models (StemGNN, AGCRN, and MTGNN) and Transformer-based models (Autoformer and Informer) on the Wiki and Traffic datasets, respectively. We use the same input length ($T=12$) and prediction length $(\tau=12)$ for the analysis in the five methods and the results are shown in Table \ref{tab:efficiency}. From the table, we can find that: 1) the Transformer-based models require less training time than GNN-based models and our model since they only model the intra-series dependencies while GNN-based models and our model consider both intra- and inter-series dependencies. 2) Compared with the GNN-based models, our proposed model performs more efficiently because the complexity of our model is $\mathcal{O}(N \times T)$ while the AGCRN and MTGNN are $\mathcal{O}(N^2)$ and StemGNN is $\mathcal{O}(N^3)$. 3) Transformer-based models consume more parameters than GNN-based models since the self-attention mechanism needs high memory usage. 4) The parameter count of our model is larger than GNN-based models because our model encodes not only timestamp-wise and variable-wise interactions but also multi-order couplings. Moreover, our model simultaneously models the intra- and inter-series dependencies while GNN-based models separately model them.

\begin{table}[]
    \centering
    \caption{Results of efficiency analysis on Wiki dataset (variables=1000, samples=803) and Traffic dataset (variables=962, samples=10560).}
    \resizebox{\textwidth}{!}{
    \begin{tabular}{l c c c c c c}
    \toprule
         & \multicolumn{6}{c}{Training time (s/epoch)} \\
         \cline{2-7}
         & DeepCN &StemGNN & AGCRN & MTGNN & Autoformer & Informer\\    
         \midrule
        Wiki & 9.61 & 92.59 & 22.48 & 27.76 & 2.39 & 2.64\\
        Traffic & 62.52 & 201.69 & 166.61 & 169.34 & 14.48& 12.99\\
    \bottomrule
    \toprule
         & \multicolumn{6}{c}{Parameters} \\
         \cline{2-7}
         & DeepCN &StemGNN & AGCRN & MTGNN & Autoformer & Informer\\    
         \midrule
        Wiki & 7.91M & 4.10M & 0.76M & 1.53M & 15.6M & 14.9M\\
        Traffic & 8.74M & 3.88M & 0.75M & 1.48M & 15.4M & 14.8M\\
    \bottomrule
    \end{tabular}}
    \label{tab:efficiency}
\end{table}

\begin{figure}[ht]
    \centering
    \subfigure[Traffic dataset]
    {
    \centering
    \includegraphics[width=2in]{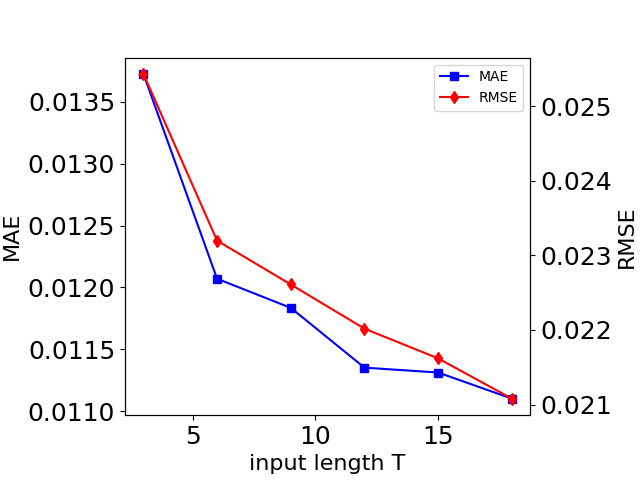}
    }
    \subfigure[ECG dataset]
    {
    \centering
    \includegraphics[width=2in]{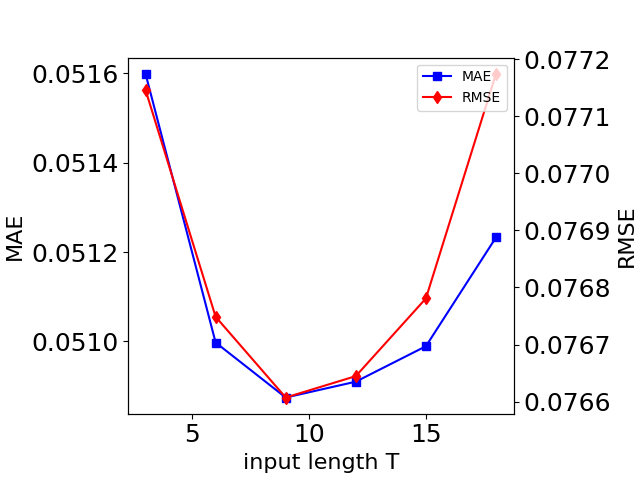}
    }
    \caption{Parameter sensitivity analysis about input length. We compare error results (MAE and RMSE) under different input lengths on the Traffic and ECG datasets, respectively.}
    \label{fig:parameter_sensitivity_input_length}
\end{figure}

\begin{figure}
    \centering
    \subfigure[Traffic dataset]
    {
    \centering
    \includegraphics[width=2in]{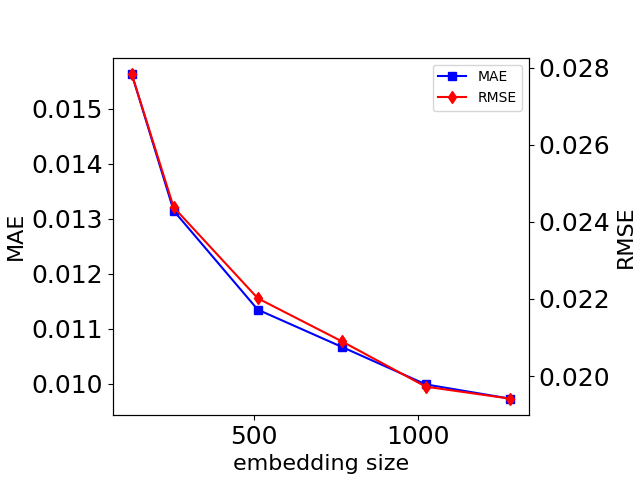}
    }
    \subfigure[ECG dataset]
    {
    \centering
    \includegraphics[width=2in]{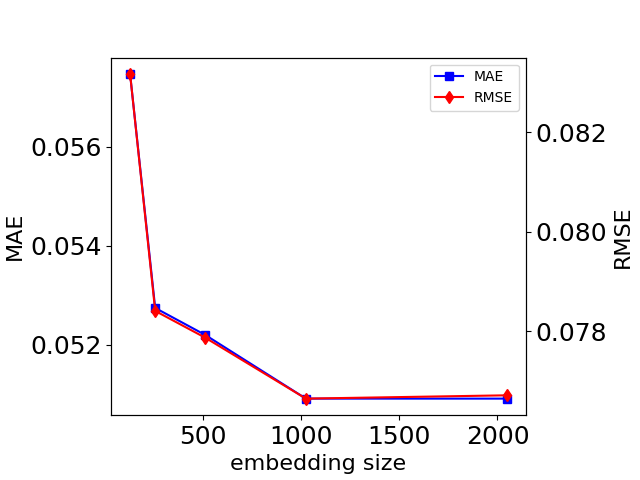}
    }
    \caption{Parameter sensitivity analysis about the embedding size. We compare error results (MAE and RMSE) under different embedding sizes on the Traffic and ECG datasets, respectively.}
    \label{fig:parameter_sensitivity_embedding_size}
\end{figure}

\textbf{Parameter Sensitivity}.\label{experiment_parameter} We perform parameter sensitivity tests of input length $T$ and embedding size $d$ on the Traffic and ECG datasets. All parameters of our model under study are held constant except the input length and the embedding size. (1) \textit{Input length}. The input length reflects the time lag effects and affects the final accuracy. We turn over it with the value $\{3, 6, 9, 12, 15, 18\}$ for the Traffic and ECG datasets, and the result is shown in Fig. \ref{fig:parameter_sensitivity_input_length}. Fig. \ref{fig:parameter_sensitivity_input_length}(a) shows that with the input length increasing, the accuracy becomes better since the long input length can bring more information and this also demonstrates the relationships between variables have time lag effects which are introduced in Section \ref{coupling_mechanism}.  Fig. \ref{fig:parameter_sensitivity_input_length}(b) shows that with the input length increasing, the performance first improves and then decreases due to data redundancy or overfitting. (2) \textit{Embedding size}. The embedding size affects the representation ability and we choose the embedding size over the set $\{128, 256, 512, 768, 1024,1280\}$ for the Traffic dataset and $\{128, 256, 512, 1024, 2048\}$ for ECG dataset. We choose different value sets for the two datasets because of the memory limit. Fig. \ref{fig:parameter_sensitivity_embedding_size}(a) demonstrates that the performance becomes better with the increase of embedding size while Fig. \ref{fig:parameter_sensitivity_embedding_size}(b) shows that the performance first improves and then keeps almost unchanged. Unlike the data on the ECG dataset, the data on the Traffic dataset exhibits strong couplings between the data. Then compared with the ECG dataset, the data on the Traffic dataset has stronger time lag effects (comparing  Fig. \ref{fig:parameter_sensitivity_input_length}(a) with Fig. \ref{fig:parameter_sensitivity_input_length}(b)), and needs bigger embedding size to represent the complicated relationships (comparing Fig. \ref{fig:parameter_sensitivity_embedding_size}(a) with Fig. \ref{fig:parameter_sensitivity_embedding_size}(b)).

\subsection{Visualization}
To gain a better understanding of the coupling mechanism for modeling intra- and inter-series relationships among time series data, we further conduct experiments on the Wiki dataset to visualize the different order couplings. 

\begin{figure*}
    \subfigure[Heatmap visualization of multi-order couplings]{
	\begin{minipage}[b]{0.57\columnwidth}
		\includegraphics[width=0.48\linewidth,trim=52 40 52 50,clip]{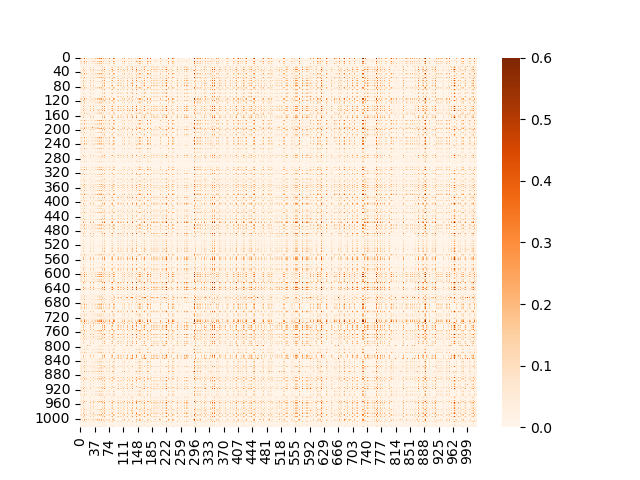}
        \includegraphics[width=0.48\linewidth,trim=52 40 52 50,clip]{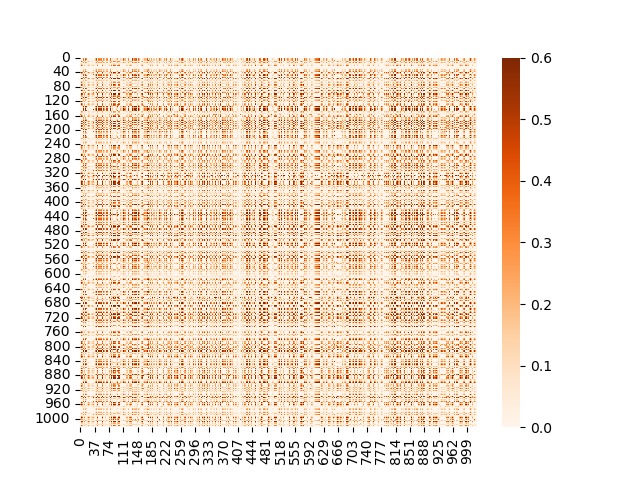}
  
        \includegraphics[width=0.48\linewidth,trim=52 40 52 50,clip]{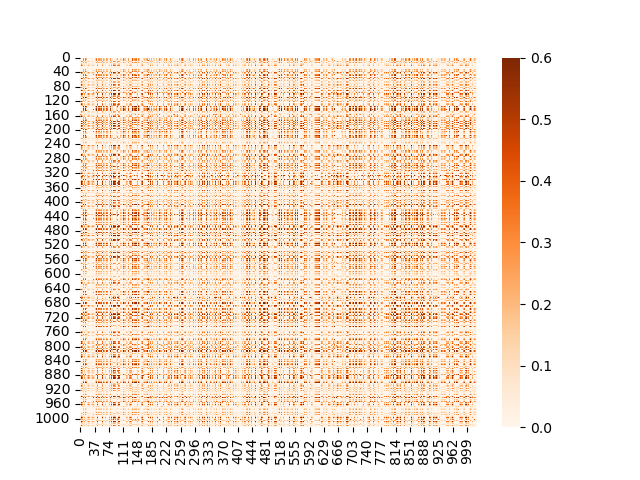}
        \includegraphics[width=0.48\linewidth,trim=52 40 52 50,clip]{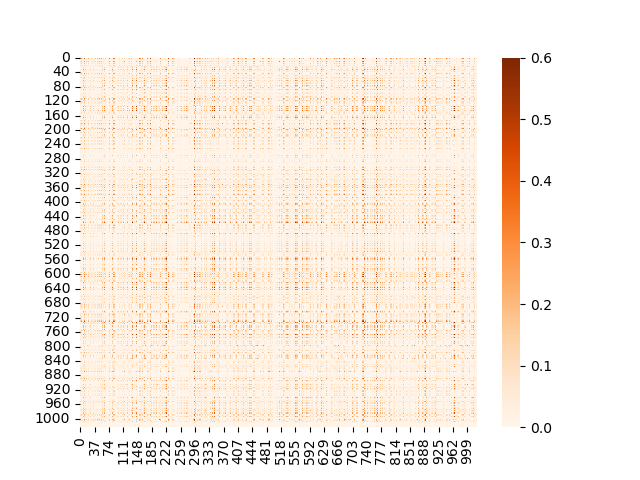}
        \label{fig:wiki_coupling}
	\end{minipage}
    }
    \subfigure[t-SNE visualization of multi-order couplings]{
	\begin{minipage}[b]{0.40\columnwidth}
		\centering
		\includegraphics[width=1\linewidth,trim=50 78 50 80,clip]{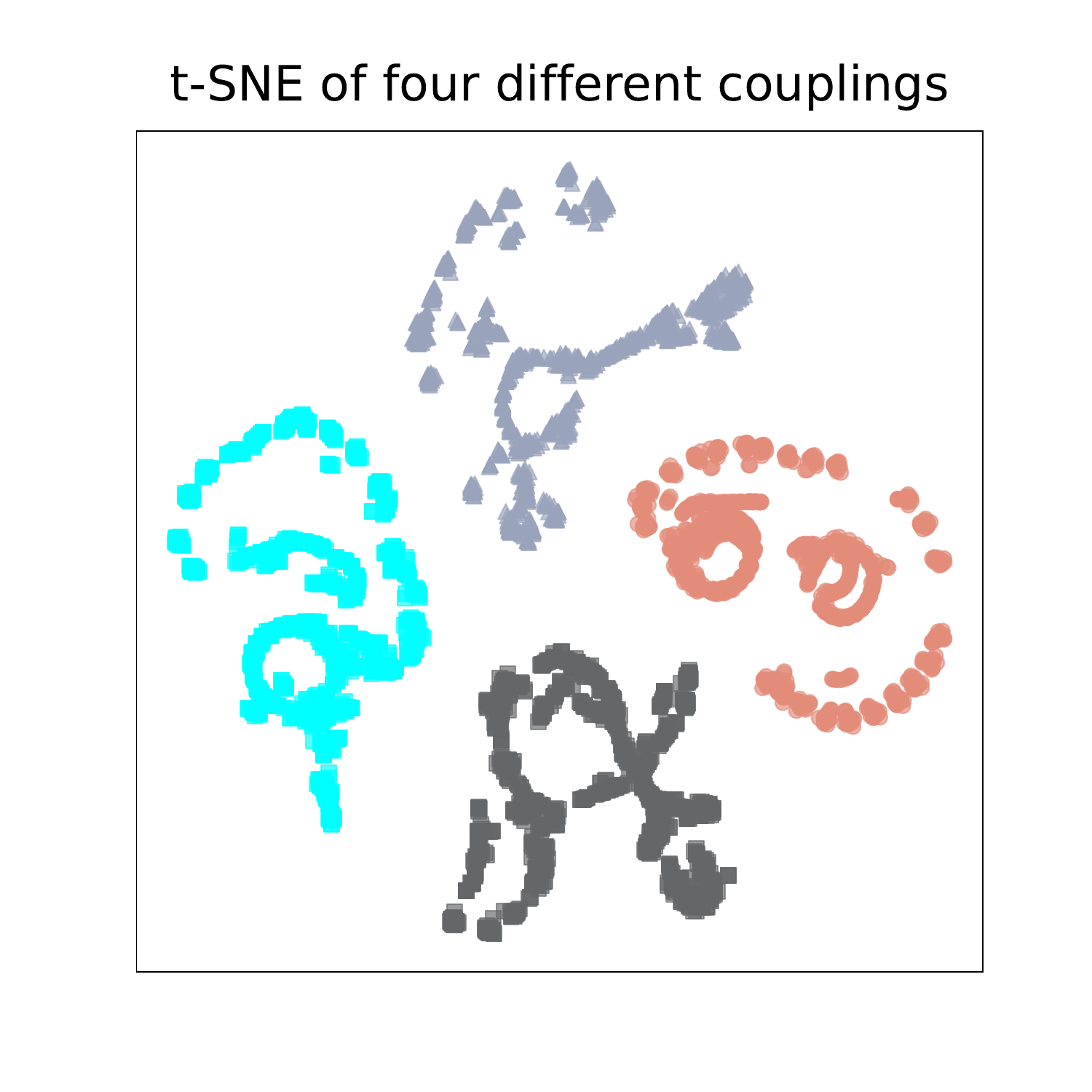}
        \label{fig:tsne}
	\end{minipage}
    }
    \caption{(a) Visualizations of multi-order couplings learned from the Wiki dataset. (b) Visualizations about t-SNE of multi-order couplings learned from the Wiki dataset. Different subfigures/colors correspond to different order couplings: specifically first-order (upper left, blue), second-order(upper right, black), third-order (bottom left, red), and fourth-order (bottom right, grey).}
\end{figure*}

\textbf{Visualization about heatmap of multi-order couplings}\quad In order to showcase the different impacts of each order of couplings in modeling the intra- and inter-series relationships, we generate the corresponding visual heatmap to represent them. Specifically, given the dense representations of couplings denoted as $\mathcal{X} \in \mathbb{R}^{\ell \times d}$, for each order $\ell$, we compute the weight matrix as follows: $W_{vis}^{(\ell)}=\mathcal{X}^{(\ell)} \times \{\mathcal{X}^{(\ell)}\}^T$. Subsequently, we visualize $W_{vis}^{(\ell)}\in \mathbb{R}^{d\times d}$ via a heatmap. In our experiments, we choose four order couplings (i.e., $\ell=4$), and the results are presented in Fig. \ref{fig:wiki_coupling}. The figure demonstrates that: (1) All four order couplings are indispensable. (2) Among the four-order couplings, the second-order and third-order couplings hold greater significance in capturing both intra- and inter-series relationships. This also explains that although the SOTA GNN- and Transformer-based models are primarily based on second-order or third-order interactions while ignoring first-order and fourth-order couplings, they have still achieved competitive results in the realm of MTS forecasting.


\textbf{Visualization about t-SNE of multi-order couplings}\quad To investigate the relation between different order couplings, we create visualizations using t-SNE for these couplings. Concretely, for the dense representations of couplings represented as $\mathcal{X} \in \mathbb{R}^{\ell \times d}$, we generate t-SNE visualizations for each order of couplings, denoted as $\mathcal{X}^{(\ell)} \in \mathbb{R}^{d}$, individually. In our experiment, the order $\ell$ is 4, and the results are depicted in Fig. \ref{fig:tsne}. The figure illustrates that the couplings of different orders are distinct and separate from each other.

Based on the insights drawn from Figs. \ref{fig:wiki_coupling} and \ref{fig:tsne}, we can conclude that the coupling mechanism comprehensively explores the different order couplings that are disentangled, and it expressively learns representations of both intra- and inter-series relationships. 


\section{Conclusion}\label{conclusion}
In this paper, to address the relationship modeling issues for MTS forecasting, we propose a novel model DeepCN for multivariate time series forecasting, including single-step forward prediction and multi-step forward prediction. Compared with the previous work, in this work, we build our model based on the couplings which can bring more comprehensive information to enhance representations of relationships among time series. Specifically, we first revisit the relationships among time series from the perspective of mutual information. Then based on the analysis, we design a coupling mechanism to learn the hierarchical and diverse couplings to represent the relationships that can exploit the intra- and inter-series relationships simultaneously. The coupling mechanism can model multi-order couplings and account for the time lag effect explicitly. After that, since different variables exhibit different patterns, we leverage a coupled variable representation module to learn the variable relationship representation. Finally, we make predictions by one forward step which can avoid error accumulation. The one-forward-step method makes our model more efficient and stable in the multi-step forward prediction. We conduct extensive experiments on seven real-world datasets and compare our model with other state-of-the-art baselines. The experimental results show that our model achieves superior performance. Furthermore, more analysis of our model demonstrates that different order couplings impose different effects on different datasets which gives us enlightenment for handling different types of MTS data. Compared with the state-of-the-art GNN-based models, DeepCN has relatively higher model complexity in terms of parameter volumes. In the future, we plan to move our work forward to reduce the parameter counts and the time complexity. 


\begin{acks}
This work was supported in part by the National Key Research and Development Program of China (Grant No. 2019YFB1406303).
\end{acks}

\bibliographystyle{ACM-Reference-Format}
\bibliography{sample-base}


\end{document}